\documentclass[lettersize,journal]{IEEEtran}
\usepackage{amsmath,amsfonts}
\usepackage{algorithmic}
\usepackage{algorithm}
\usepackage{array}
\usepackage[caption=false]{subfig}

\usepackage{textcomp}
\usepackage{stfloats}
\usepackage{url}
\usepackage{verbatim}
\usepackage{graphicx}
\usepackage{cite}
\usepackage[inline]{enumitem}

\usepackage{booktabs}
\usepackage{threeparttable}
\usepackage{multirow}
\usepackage{colortbl}
\usepackage[dvipsnames,svgnames]{xcolor}
\usepackage{tabularx}

\usepackage[colorlinks,linkcolor=blue]{hyperref}

\begin{document}
\definecolor{newtext_blue}{HTML}{1E6FA9} 
\definecolor{revised_red}{HTML}{C33390} 
\newcommand{\newtext}[1]{\textcolor{newtext_blue}{#1}}   
\newcommand{\revised}[1]{\textcolor{revised_red}{#1}}    

\title{Disentangling Bias by Modeling Intra- and Inter-modal Causal Attention for Multimodal Sentiment Analysis}

\author{Menghua Jiang, Yuxia Lin, Baoliang Chen, Haifeng Hu, Yuncheng Jiang, and Sijie Mai
\thanks{This work is supported by the Guangdong Philosophy and Social Sciences Planning Project (No. GD26YJY34). (\textit{Corresponding author: Sijie Mai.})}

\thanks{Menghua Jiang, Yuxia Lin, Baoliang Chen, Yuncheng Jiang and Sijie Mai are with the School of Computer Science, South China Normal University, Guangzhou 510631, China; Haifeng Hu is with the School of Electronics and Information Technology, Sun Yat-sen University, Guangzhou 510275, China. E-mail: \{jiangmenghua, 2025023399, jiangyuncheng, sijiemai\}@m.scnu.edu.cn; blchen6-c@my.cityu.edu.hk; huhaif@mail.sysu.edu.cn.}
}





\markboth{Journal of \LaTeX\ Class Files,~Vol.~14, No.~8, August~2021}%
{Shell \MakeLowercase{\textit{et al.}}: A Sample Article Using IEEEtran.cls for IEEE Journals}


\maketitle

\begin{abstract}
Multimodal sentiment analysis (MSA) aims to understand human emotions by integrating information from multiple modalities, such as text, audio, and visual data. However, existing methods often suffer from spurious correlations both within and across modalities, leading models to rely on statistical shortcuts rather than true causal relationships, thereby undermining generalization. To mitigate this issue, we propose a Multi-relational Multimodal Causal Intervention (MMCI) framework, which leverages the backdoor adjustment from causal theory to address the confounding effects of such shortcuts. Specifically, we first model the multimodal inputs as a multi-relational graph to explicitly capture intra- and inter-modal dependencies. Then, we apply an attention mechanism to separately estimate and disentangle the causal features and shortcut features corresponding to these intra- and inter-modal relations. Finally, by approximating backdoor adjustment, we stratify the shortcut features and dynamically combine them with the causal features to encourage MMCI to produce stable predictions under distribution shifts. Extensive experiments on several standard MSA datasets and out-of-distribution (OOD) settings demonstrate that our method effectively suppresses biases and improves performance.
\end{abstract}


\section{Introduction}
With the rapid development of social media and the internet, multimodal data has proliferated in online environments, offering users richer ways to express their emotions and opinions. Meanwhile, this trend has increased the demand for more accurate sentiment analysis techniques, shifting the research paradigm from unimodal text-based sentiment analysis to multimodal sentiment analysis (MSA) \cite{gandhi2023multimodal}. By leveraging information from text, audio, and visual modalities, MSA enables a more comprehensive understanding of user emotions.

\begin{figure}[htbp]
    \centering
    \includegraphics[width=\linewidth]{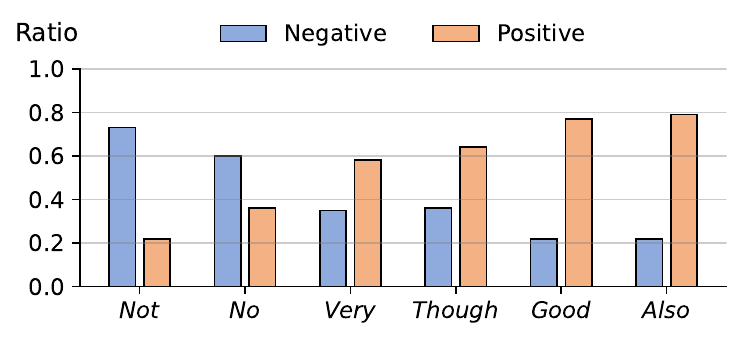} 
    \caption{The distribution of several words from the training set on the CMU-MOSI dataset~\cite{CMU-MOSI2016}.}
    \label{fig:words_count}
\end{figure}

\begin{figure}[htbp]
    \centering
    \includegraphics[width=\linewidth]{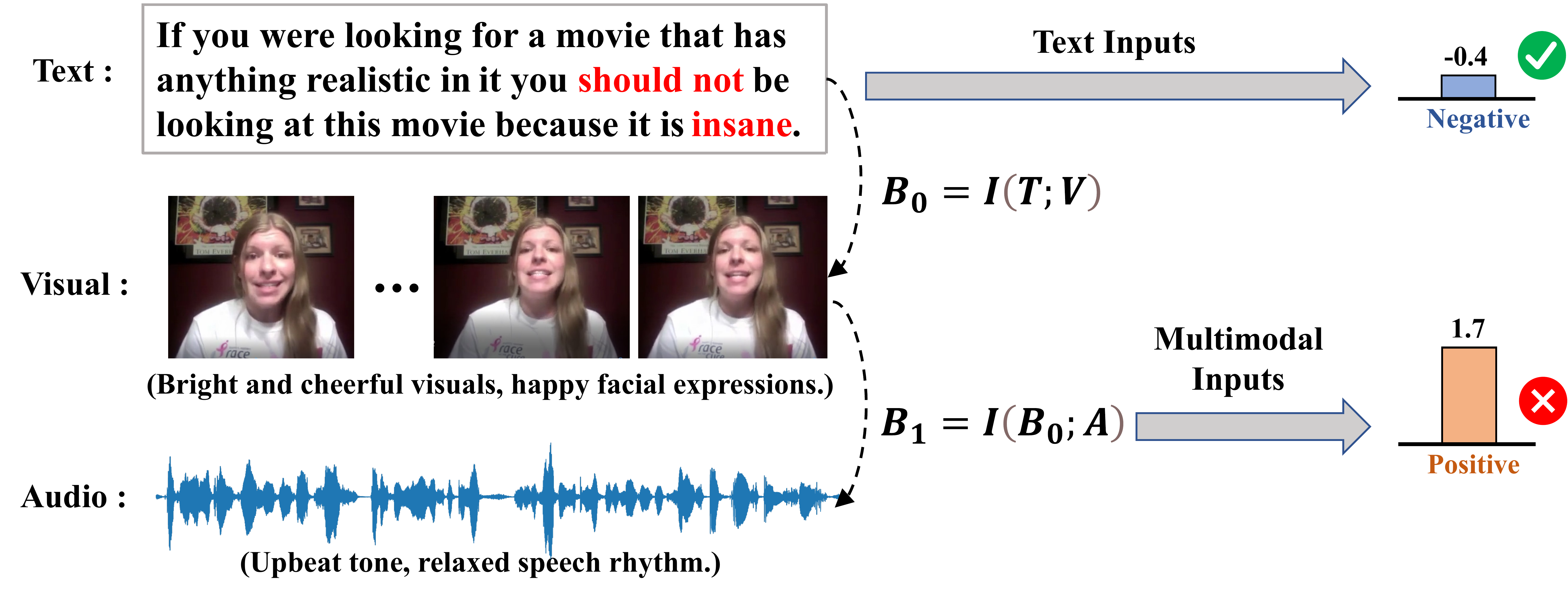} 
    \caption{A testing sample from the CMU-MOSI \cite{CMU-MOSI2016} dataset has a sentiment label of -2.4. The re-implemented ITHP model \cite{ITHP2024} makes correct predictions using text inputs but fails on multimodal inputs.}
    \label{fig:intro}
\end{figure}

The main challenge of MSA lies in effectively integrating heterogeneous modality information to achieve accurate sentiment prediction. However, due to the high-dimensional nature and complex interdependencies of multimodal data, it remains significantly difficult to accurately model the interactions among modalities \cite{zhu2023multimodal}. To address this challenge, advanced modeling techniques are required to enable multimodal fusion and fully exploit cross-modal interactions. Existing studies have proposed various fusion approaches, such as attention-based cross-modal interaction \cite{MulT2019,MAG2020}, adversarial learning \cite{mai2020modality}, and methods based on the information bottleneck principle \cite{MIB2023,ITHP2024}. Although these methods have achieved impressive performance on benchmark datasets, they tend to overly focus on maximizing inter-modality correlation and often overlook the issue of spurious correlations \cite{CCIM2023,MulDeF2024,CCIM2024-tpami}. As a result, models may learn non-causal statistical patterns within or across modalities rather than capturing causal relationships.

Spurious correlations in MSA mainly manifest in two forms: intra-modal bias and inter-modal bias. The first is intra-modal bias, where models rely on superficial features within a single modality instead of genuine sentiment-related semantics. For example, certain high-frequency words often co-occur with a particular sentiment label in the training data. Consequently, models tend to exploit these lexical shortcuts while overlooking richer contextual semantic information \cite{MCIS2024,CLUE2022}. As illustrated in Fig. \ref{fig:words_count}, the word ``Also" frequently appears together with positive sentiment samples during training. Rather than learning its contextual meaning, the model mistakenly treats this word as a predictive cue. Consequently, the model continues to rely on this shortcut during testing, causing samples containing ``Also" to be incorrectly classified as expressing positive sentiment. The second is inter-modal bias, where spurious correlations simultaneously exist across multiple modalities and are further amplified during multimodal fusion. For instance, background colors or lighting conditions in the visual modality may coincidentally correlate with particular sentiment labels, while background music in the audio modality may be incorrectly interpreted as emotional information from the speaker \cite{GEAR2023}. Since these irrelevant patterns exist across multiple modalities, multimodal fusion models may reinforce them instead of extracting genuine emotional cues. Fig. \ref{fig:intro} presents a representative example. ITHP \cite{ITHP2024}, which explicitly maximizes mutual information among modalities based on the information bottleneck principle, inadvertently reinforces these cross-modal spurious correlations during feature fusion, leading to biased sentiment predictions. In contrast, using only the textual modality avoids such misleading cross-modal interference and therefore yields a more accurate prediction.

Unlike traditional machine learning models that rely on statistical correlations from observational data, human decision-making fundamentally follows a causal inference mechanism. For example, when medical institutions study the relationship between smoking and lung cancer, a traditional model may conclude that “smokers have a higher incidence of lung cancer” based purely on correlation. However, such a model cannot determine whether this reflects a true causal effect or a spurious association caused by confounders such as age or occupational exposure. In contrast, human researchers typically use stratification or adjustment to control key confounders and identify potential backdoor paths (e.g., smoking $\leftarrow$ occupation $\rightarrow$ lung cancer). This allows researchers to uncover true causal relationships even from biased observations. Similarly, in multimodal learning, intervention mechanisms based on structured causal models should be established to block spurious association paths.

Based on these observations, we propose the Multi-relational Multimodal Causal Intervention (MMCI) framework to alleviate the adverse effects of spurious correlations. Unlike existing methods that either rely on predefined bias assumptions or are tailored to individual modalities or specific modality combinations, MMCI provides a unified causal framework that simultaneously alleviates both intra-modal and inter-modal confounders. By avoiding reliance on particular bias patterns, MMCI generalizes better across diverse distribution shifts. Specifically, we first construct a causal graph to analyze the underlying mechanism of spurious correlations in multimodal fusion. Guided by this causal graph, multimodal inputs are represented as a multi-relational graph that explicitly models both intra-modal and inter-modal dependencies. Based on this graph representation, Graph Attention Networks (GATs) are employed to capture informative interactions and distinguish causal features from shortcut features within and across modalities. Furthermore, backdoor adjustment is applied to shortcut features by dynamically stratifying and recombining them with causal features, thereby suppressing the influence of spurious correlations during optimization. Consequently, MMCI encourages the causal branch to learn more stable representations while reducing reliance on unstable shortcut patterns.
Extensive experiments on multiple benchmark datasets demonstrate that MMCI consistently achieves state-of-the-art performance, yielding more reliable sentiment predictions while exhibiting superior robustness in out-of-distribution (OOD) settings.

Our main contributions can be summarized as follows:
\begin{itemize}
\item We construct a tailored causal graph to analyze spurious correlations arising during heterogeneous modality fusion in MSA, formalizing them as confounders that can mislead the model toward biased predictions.
\item By leveraging backdoor adjustment, we propose MMCI, a framework that encourages the model to capture causal features while reducing reliance on spurious shortcuts.
\item Extensive experiments on MSA datasets including CMU-MOSI, CMU-MOSEI, and CH-SIMS, as well as under OOD settings, demonstrate the effectiveness of MMCI. Furthermore, in-depth analyses highlight the plausibility of our method.
\end{itemize}

\section{Related Work}

\subsection{Multimodal Sentiment Analysis}
In recent years, most studies on MSA have focused on developing advanced fusion techniques to learn more discriminative multimodal representations \cite{zhu2023multimodal}. Early approaches primarily adopted feature-level fusion, such as Tensor Fusion Networks (TFN) \cite{TFN2017}, to explicitly model joint interactions among different modalities. Subsequently, decision-level fusion methods combined predictions from individual modalities to improve prediction robustness \cite{poria2016convolutional,yu2019adapting}. With the remarkable success of attention mechanisms~\cite{vaswani2017attention,ahmad2025prost,ahmad2026misinformation}, Transformer-based architectures have become the dominant paradigm for multimodal representation learning \cite{yang2020cm,MulT2019,BBFN2021,TETFN2023,DecAlign2026}. Representative methods enhance cross-modal interactions through multimodal pre-training objectives \cite{MAG2020,ALMT2023,CENet2022} or ensemble learning strategies \cite{liu2024ensemble}. For example, AOBERT \cite{AOBERT2023} introduces multimodal masked language modeling and alignment prediction objectives to further pre-train BERT~\cite{BERT2019}, thereby strengthening both intra-modal and inter-modal semantic correlations. 
Beyond feature fusion, another line of research aims to learn more informative multimodal representations through representation disentanglement. These methods decompose multimodal features into modality-shared and modality-specific components using dedicated encoders \cite{MISA2020,ConFEDE2023,GLoMo2024}. To further improve cross-modal representation learning, recent studies employ various techniques, including canonical correlation analysis~\cite{sun2020learning}, Kullback--Leibler divergence~\cite{shankar2022multimodal}, graph-based approaches~\cite{ahmad2025hierarchical,HGraph-CL2022,MTAG2021}, and the information bottleneck~\cite{MIB2023,OMIB2025}, to reduce distribution discrepancies across modalities and suppress redundant information. For instance, ITHP \cite{ITHP2024} leverages the information bottleneck to identify a primary modality while treating the remaining modalities as auxiliary detectors to facilitate information distillation.

However, despite strong performance, these methods mainly focus on modeling cross-modal correlations and inevitably capture spurious correlations caused by dataset biases or optimization objectives, leading to biased predictions and poor OOD generalization. In contrast, MMCI explicitly models the causal generation process and applies causal intervention to disentangle causal and shortcut features, improving robustness.

\subsection{Causal Inference in Multimodal Learning}

Recently, causal inference has attracted increasing attention in deep learning owing to its ability to uncover causal effects among observed variables rather than merely modeling superficial correlations, thereby improving model generalization in complex data \cite{zhang2021multi,hong2021unbiased,qian2021counterfactual,wang2022causal}. Consequently, it has achieved remarkable success in a variety of applications, including visual question answering~\cite{liu2023cross}, recommender systems~\cite{wang2022causal}, and graph classification~\cite{CAL2022,DIR2022}.

In multimodal learning, many causal methods have been proposed to mitigate data biases, which can be broadly categorized into two main directions: counterfactual reasoning and causal intervention~\cite{pearl2009causality}. Counterfactual reasoning focuses on hypothetical scenarios by examining how observed variables would behave under alternative interventions. For example, Sun et al. designed a counterfactual framework that captures the direct effect of the textual modality through an additional text model, thereby mitigating spurious correlations between textual words and sentiment labels~\cite{CLUE2022}. Niu et al. modeled language bias as the direct causal effect of questions on answers and reduced such bias by subtracting the direct language effect from the total causal effect, achieving remarkable success in visual question answering~\cite{niu2021counterfactual}. Yang et al. proposed a multimodal counterfactual inference framework for sentiment analysis, which constructs two counterfactual scenarios to purify and alleviate multimodal utterance-level label bias and word-level context bias~\cite{MCIS2024}. Huang et al. introduced a counterfactual cross-modal attention mechanism to generate more reliable attention distributions and incorporate causal information into modality fusion, thereby improving the quality of multimodal aggregation~\cite{AtCAF2025}. From a complementary perspective of actual causality, Chen et al. analyzed modal preemption in MSA, leveraging encoder gradient norms to estimate modal preemption and dynamically balance the optimization of different modalities~\cite{AGS-SMoE2025}.

Causal intervention aims to actively manipulate independent variables to break the natural dependencies among variables, thereby mitigating the influence of confounders~\cite{zhang2021causal}. Yang et al. proposed a contextual causal intervention module based on backdoor adjustment to approximate causal effects during model training, thereby alleviating context bias in the dataset~\cite{CCIM2023,CCIM2024-tpami}. Sun et al. developed a general debiasing framework based on inverse probability weighting, which adaptively assigns smaller weights to samples exhibiting more severe spurious correlations, thus encouraging robust feature learning for sentiment prediction~\cite{GEAR2023}. Xu et al. proposed a subject causal intervention module, which performs causal intervention on subject-related biases through subject-specific feature disentanglement and subject prototype-based confounder construction, thereby reducing the misleading influence of inter-subject semantic discrepancies~\cite{SuCI2025}. Huan et al. further introduced a multimodal debiasing framework that alleviates label bias via front-door causal adjustment during training and removes multimodal bias through counterfactual reasoning during inference, thus improving the generalization ability of MSA models under OOD settings~\cite{MulDeF2024}. Liu et al. proposed a plug-and-play feature causality decomposition method that separates multimodal representations into modality-invariant, modality-specific, and redundant components, and removes uncertainty-related redundancy through backdoor adjustment~\cite{FCD2025}. Li et al. modeled class-wise co-occurring biases as learnable contextual prototypes and performed adaptive causal intervention for multi-label image classification using only image-level supervision~\cite{ProCI2026}.

However, most of the aforementioned methods either focus on a single modality or specific modality pairs, rely on explicitly annotated bias types, or are tailored to particular causal issues or task settings, which often require domain knowledge~\cite{nam2020learning}. This limits their broader applicability to complex multimodal data, where biases are usually implicit rather than predefined, may arise from multiple sources, and can be further amplified during multimodal fusion. In contrast, MMCI provides a general causal intervention framework for both intra-modal and inter-modal biases without requiring predefined bias types, thereby substantially improving generalization in complex multimodal scenarios.

\begin{table}[t]
\centering
\setlength{\belowcaptionskip}{4pt}
\renewcommand\arraystretch{1.1}
\caption{Summary of notations.}
\label{tab:notation}
\resizebox{\linewidth}{!}{
\begin{tabular}{ll}
\toprule
\textbf{Notation} & \textbf{Definition} \\
\midrule
$L_m$, $d_m$
& Sequence length and feature dimension of modality $m$. \\
$y$, $\hat{y}$
& Ground-truth label and predicted sentiment value. \\
$\mathbb{D}$, $N$
& Training set and number of training samples. \\
$\mathcal{Z}$
& Value space of the confounding variable $Z$. \\
\( X_m \in \mathbb{R}^{L \times d} \)
& Aligned feature representation of modality $m$. \\
$H_m \in \mathbb{R}^{L \times d}$
& Encoded representation of modality $m$. \\
$L$, $d$
& Shared sequence length and embedding dimension. \\

$\mathcal{G}=(\mathcal{V},\mathcal{E})$
& Constructed multi-relational multimodal graph. \\
$\mathcal{V}$
& Node set of graph $\mathcal{G}$. \\
$\mathcal{E}$
& Multi-relational adjacency tensor of graph $\mathcal{G}$. \\
$\mathcal{R}$
& Set of relation types. \\
$\mathcal{E}^{(r)}$
& Adjacency matrix of relation type $r$. \\
$\alpha_{ij,c}^{(r)}$, $\alpha_{ij,s}^{(r)}$
& Causal and shortcut attention weights for edge $(i,j)$. \\
$H_c^{(r)}$, $H_s^{(r)}$
& Causal and shortcut node representations under relation $r$. \\
$H_c$, $H_s$
& Aggregated causal and shortcut node representations. \\
$\Phi_c$, $\Phi_s$
& Predictors for the causal and shortcut branches. \\
$\hat{K}$, $H_s^{(k)}$
& Shortcut memory bank and its $k$-th sampled representation. \\

\bottomrule
\end{tabular}
}
\end{table}

\section{Methodology}
In this section, we first examine modality fusion in MSA from a causal perspective. Based on our causal hypothesis, we analyze how intra-modal and inter-modal confounding effects give rise to spurious correlations and further impair model generalization. To mitigate these effects, we introduce the proposed MMCI framework, which performs causal intervention on both intra-modal and inter-modal biases. For clarity, the main mathematical notations used throughout this section are summarized in Table~\ref{tab:notation}.

\subsection{Task Formulation}
MSA aims to predict the sentiment state from text (\( t \)), audio (\( a \)), and visual (\( v \)) modalities extracted from videos. Each modality is represented as a sequence, i.e., \( t \in \mathbb{R}^{L_t \times d_t} \), \( a \in \mathbb{R}^{L_a \times d_a} \), and \( v \in \mathbb{R}^{L_v \times d_v} \), where \( L_t, L_a, L_v \) denote the sequence lengths and \( d_t, d_a, d_v \) denote the feature dimensions of the respective modalities. Given a training set \( \mathbb{D} = \{(t_{i}, a_{i}, v_{i}, y_i)\}_{i=1}^N \), where each sample contains three modalities and a sentiment label, the goal is to learn a model \( \mathcal{M}_\phi \) that predicts \( \hat{y}_i = \mathcal{M}_\phi(t_{i}, a_{i}, v_{i}) \), where \( \phi \) represents the learnable parameters.

\begin{figure}[t]
    \centering
    \includegraphics[width=\linewidth]{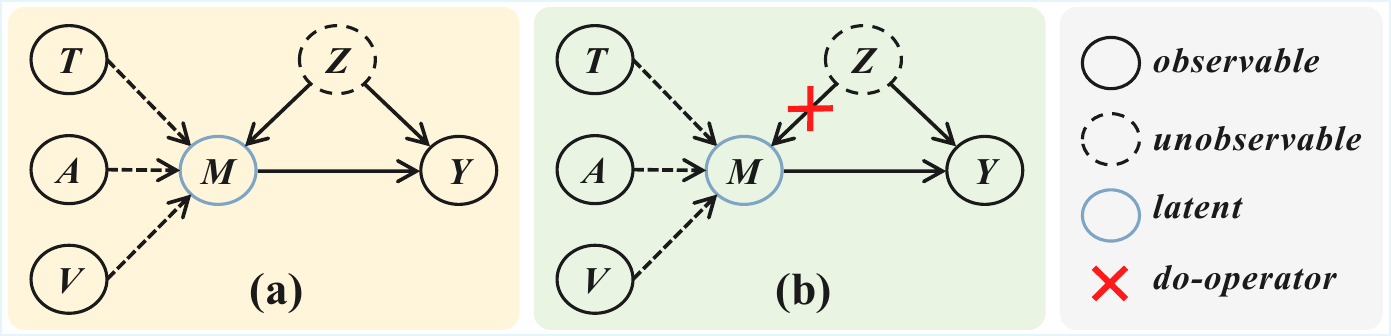}
    \caption{(a) A causal graph tailored for modality fusion in MSA. (b) The same graph with backdoor adjustment to address confounding effects.}
    \label{fig:2}
\end{figure}

\subsection{A Causal Perspective on MSA}
\label{A Causal Perspective on MSA}
To clarify how spurious correlations arise and propagate during multimodal fusion, we formulate the learning process of MSA models from a causal perspective, as illustrated in Fig.~\ref{fig:2} (a). Under our causal hypothesis, shortcut features are regarded as confounders that may bias the learned multimodal representations and sentiment predictions. The causal graph contains the observable textual ($T$), acoustic ($A$), and visual ($V$) inputs, the learned multimodal representation ($M$), the sentiment prediction ($Y$), and an unobserved confounding variable ($Z$). Here, $Z$ does not denote a single predefined bias attribute. Instead, it summarizes unobserved nuisance factors in the data-generation process, such as high-frequency words, speaker identity, topics, environmental background colors, and other dataset-specific biases. In this graph, a directed edge from one variable to another indicates a causal dependency. We provide a detailed explanation of the causal graph as follows.

\begin{itemize}
    \item \textbf{\textit{Modality Fusion}: $\{T,A,V\}\rightarrow M$.}
    An MSA model integrates textual, acoustic, and visual information into a discriminative multimodal representation $M$. For example, negative utterances, frustrated vocal tones, and frowning expressions can jointly provide stronger evidence of negative sentiment than any single modality.

    \item \textbf{\textit{Confounding Bias}: $M\leftarrow Z\rightarrow Y$.} The variable $Z$ represents unobserved nuisance factors affecting both $M$ and the label distribution $Y$. For intra-modal bias, the word ``Also'' may frequently co-occur with positive labels and thus be learned as a positive shortcut. For inter-modal bias, background colors or lighting conditions may coincidentally correlate with certain sentiment polarities and be reinforced by acoustic cues such as background music. These dataset-specific biases may become spuriously associated with $Y$.

    \item \textbf{\textit{Task Prediction}: $M\rightarrow Y$.}
    The model uses $M$ to predict $Y$. Ideally, it relies on genuine sentiment cues such as semantic content, vocal tone, and facial expressions. However, shortcuts such as ``Also'' or particular background colors may cause an incorrect positive prediction even when the genuine cues indicate otherwise.
\end{itemize}

Taken together, the observed association between $M$ and $Y$ contains both the desired task-relevant pathway $M\rightarrow Y$ and the backdoor pathway $M\leftarrow Z\rightarrow Y$. A model trained primarily for predictive performance may exploit the latter when shortcut features are strongly correlated with sentiment labels in the training data. However, when dataset-specific bias distributions change, these correlations may no longer hold, resulting in degraded OOD generalization.

\subsection{Backdoor Adjustment}
To block the backdoor path, we introduce a causal intervention mechanism guided by backdoor adjustment~\cite{pearl2009causality}. Specifically, we replace the traditional conditional probability $P(Y \mid M)$ with the interventional probability $P(Y \mid do(M))$. Our objective is to break the spurious dependency between the multimodal representation $M$ and the confounding shortcut feature $Z$. The symbol $do(M)$ denotes the $do$-operator \cite{pearl2009causality}, which mathematically simulates a physical intervention by forcefully setting the value of $M$ and blocking its natural associations with its causes. As illustrated in Fig.~\ref{fig:2} (b), the ideal intervention cuts off the incoming edges to \(M\) (the path \(Z \rightarrow M\)), providing a theoretical target for estimating the causal effect of the fused features on sentiment prediction. Since \(Z\) is unobservable in practice, MMCI approximates this target using learned shortcut representations. According to the backdoor adjustment formula, the interventional distribution can be expanded as:
\begin{align}
    P(Y\,|\,do(M)) 
    &= \sum_{z \in \mathcal{Z}} P(Y\,|\,do(M), z) \cdot P(z\,|\,do(M)).
    \label{eq:1}
\end{align}

According to causal theory~\cite{pearl2009causality}, when $Z$ satisfies the backdoor criterion, it has two key properties that simplify Eq.~\eqref{eq:1}: 
\textbf{i)} The intervention on $M$ does not change the marginal distribution of the confounder $Z$, since cutting the backdoor path only removes the causal dependence from $Z$ to $M$ but does not alter the distribution of shortcut factors, i.e., $P(z|do(M)) = P(z)$;
\textbf{ii)} After conditioning on $Z$, the $do$-operator on $M$ becomes equivalent to observing $M$, since all non-causal paths from $M$ to $Y$ are blocked. In other words, the response of $Y$ to the causal representation can be estimated independently of the spurious dependence between causal and shortcut features, i.e., $P(Y|do(M), z) = P(Y|M, z)$.
By substituting these properties into Eq.~\eqref{eq:1}, we derive the classic backdoor adjustment formula:
\begin{align}
P(Y|do(M))
&= \sum_{z \in \mathcal{Z}} P(Y|M, z) \cdot P(z),
\label{eq:2}
\end{align}
where $\mathcal{Z}$ denotes the value space of the confounding variable $Z$, and $P(z)$ represents its prior distribution. This formulation computes the expectation of the conditional probability $P(Y \mid M, z)$ over the confounding states, effectively shielding the prediction $Y$ from the spurious correlations propagated by $Z$. 

However, implementing backdoor adjustment in MSA poses two practical challenges:
\textbf{i)} The confounders in multimodal scenarios are usually unobservable and not explicitly predefined, since they are implicitly embedded in heterogeneous modalities and may arise from multiple sources, such as high-frequency words, acoustic patterns, visual backgrounds, or their cross-modal co-occurrences;
\textbf{ii)} The high-dimensional and complex nature of $\mathcal{Z}$ makes the exhaustive summation in Eq.~\eqref{eq:2} computationally prohibitive.
To address these challenges, we propose a concise yet effective solution to approximate this adjustment efficiently.

\begin{figure*}[htbp]
    \centering
    \includegraphics[width=\linewidth]{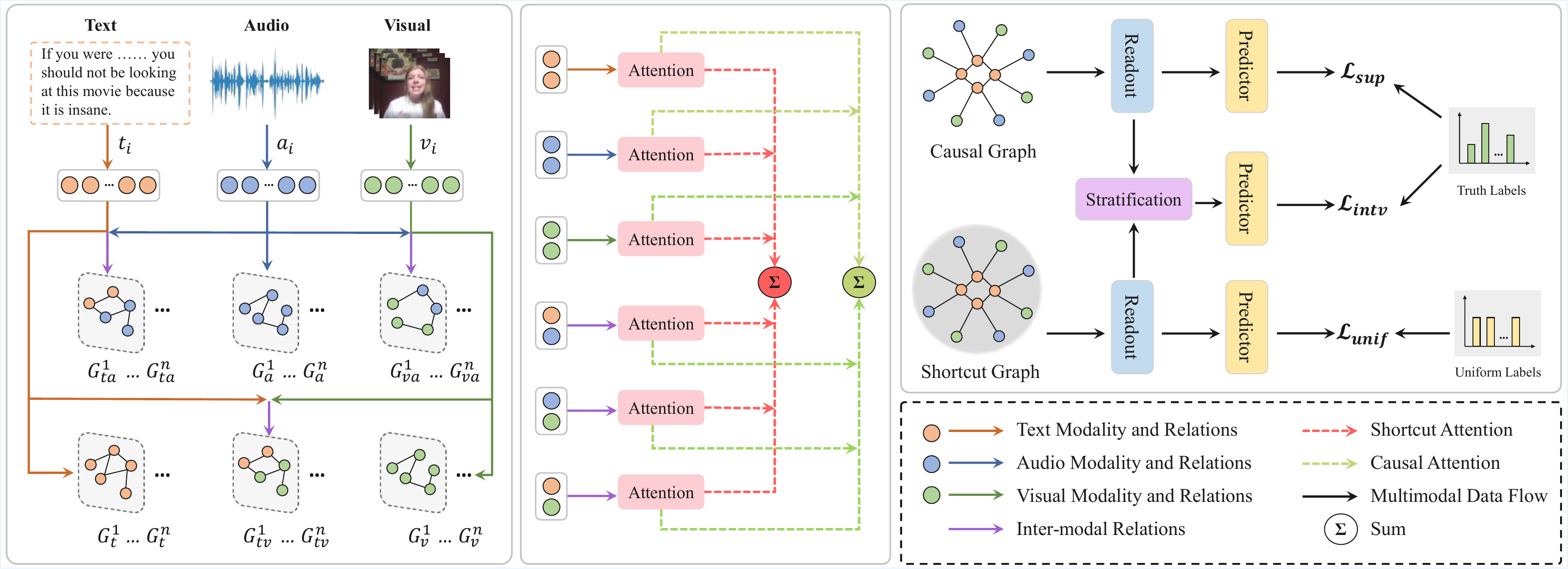}
    \caption{The illustration of the proposed MMCI consists of three main components, shown from left to right: (1) Multi-relational Graph Construction, (2) Causal and Shortcut Attention Estimation, and (3) Disentanglement and Causal Intervention.}
    \label{fig:3}
\end{figure*}

\subsection{Debiasing via MMCI}  
To translate the theoretical intervention in Eq.~\eqref{eq:2} into a practical modeling framework, we propose MMCI, as illustrated in Fig.~\ref{fig:3}. The framework consists of a preprocessing module followed by three main modules, which are described in detail below.

\subsubsection{Preprocessing} 
To facilitate the construction of the multimodal graph, we first perform temporal alignment and feature projection on the input modalities. Following prior work~\cite{MIB2023,GLoMo2024,ITHP2024}, we employ a 1D temporal convolutional layer to align all modalities to a shared sequence length \(L\) and embedding dimension \(d\). After preprocessing, the aligned features for each modality are denoted as \( X_m \in \mathbb{R}^{L \times d} \), where \( m \in \{t, a, v\} \). Each modality is then encoded by a lightweight encoder to obtain modality-specific representations: $H_m = E_m(X_m), \quad m \in \{t, a, v\},$ where \( H_m \in \mathbb{R}^{L \times d} \) denotes the encoded representation of modality \(m\). These representations are subsequently used as node features for constructing the multi-relational graph.

\subsubsection{Multi-relational Graph Construction}
We model multimodal inputs as a multi-relational graph \( \mathcal{G} = (\mathcal{V}, \mathcal{E}) \), where \(\mathcal{V} = \{h_i\}_{i=1}^{|\mathcal{V}|}\) denotes the set of node representations, with \(h_i \in \mathbb{R}^{d}\) representing the feature embedding of node \(i\), and \(|\mathcal{V}| = 3L\) denoting the total number of nodes (i.e., \(L\) nodes for each of the three modalities). The adjacency tensor \( \mathcal{E} \in \{0,1\}^{|\mathcal{V}| \times |\mathcal{V}| \times |\mathcal{R}|}\) encodes \( |\mathcal{R}|=6 \) relation types, including three intra-modal relations (text-text, visual-visual, audio-audio) and three inter-modal relations (text-visual, text-audio, visual-audio). Specifically, each relation is represented by an adjacency matrix defined as:
\begin{equation}
\mathcal{E}_{ij}^{(r)} =
\begin{cases}
1, & (i,j) \in \mathcal{R}_r, \\
0, & \text{otherwise},
\end{cases}
\label{eq:4}
\end{equation}
where \(\mathcal{R}_r\) denotes the set of node pairs \((i,j)\) connected under the \(r\)-th relation type.

For intra-modal relations, the textual modality follows \cite{HGraph-CL2022}, where the graph is constructed based on the sentence-level dependency tree. The dependency tree is obtained using spaCy~\cite{honnibal2017spacy}, and edges are established between word nodes if a syntactic dependency exists. For the visual and audio modalities, nodes correspond to local features extracted from video frames or audio segments, and edges are constructed between adjacent time steps to capture temporal continuity within each modality. 
For inter-modal relations, we adopt a temporal alignment strategy to connect heterogeneous modalities. Specifically, nodes from text, visual, and audio streams occurring at the same time step are linked to model their synchronous interactions, enabling the graph to capture cross-modal dependencies in a unified manner.

\subsubsection{Causal and Shortcut Attention Estimation}
Based on the constructed multi-relational graph $\mathcal{G}$, we employ a graph attention network (GAT) to refine node representations by aggregating neighbor information via relation-specific attention. Specifically, for each connected node pair $(i, j)$ under relation $r$, we compute a pair of complementary attention scores to distinguish between causal and shortcut influences. The features of nodes $i$ and $j$ are concatenated and fed into a relation-specific multilayer perceptron (MLP) followed by a softmax layer:
\begin{equation}
[\alpha_{ij,c}^{(r)}, \alpha_{ij,s}^{(r)}] =
\mathrm{softmax}(\mathrm{MLP}^{(r)}([h_i \Vert h_j])),
\end{equation}
where $\Vert$ denotes vector concatenation. These scores satisfy $\alpha_{ij,c}^{(r)} + \alpha_{ij,s}^{(r)} = 1$, and quantify the relative importance of causal and shortcut paths, respectively. This formulation effectively induces two distinct edge-weighted computational graphs corresponding to causal and shortcut dependencies.

Based on the learned attention scores, we construct weighted adjacency matrices for the causal and shortcut views, respectively, and update the node representations accordingly. For each relation type $r$, the relation-specific node features are updated via a message-passing mechanism:
\begin{equation}
\begin{aligned}
h_{i,c}^{(r)} &= \sigma \left( \sum_{j \in \mathcal{N}_i^{(r)}} \alpha_{ij,c}^{(r)} \mathbf{W}_c^{(r)} h_j \right), \\ 
h_{i,s}^{(r)} &= \sigma \left( \sum_{j \in \mathcal{N}_i^{(r)}} \alpha_{ij,s}^{(r)} \mathbf{W}_s^{(r)} h_j \right),
\end{aligned}
\label{eq:node_update}
\end{equation}
where $\mathbf{W}_c^{(r)}$ and $\mathbf{W}_s^{(r)}$ are trainable transformation matrices tailored to the $r$-th relation, and $\sigma$ denotes a sigmoid function that introduces non-linearity. 
Here, $\mathcal{N}_i^{(r)} = \{ j \mid \mathcal{E}_{ij}^{(r)} = 1 \}$ denotes the set of neighboring nodes of node $i$ under relation type $r$. 
To capture the holistic structural information across all relation types, we aggregate the relation-specific embeddings to derive the final causal and shortcut node representations:
\begin{equation}
H_c = \sum_{r=1}^R H_c^{(r)}, \quad H_s = \sum_{r=1}^R H_s^{(r)},
\end{equation}
where $H_c^{(r)} = [h_{1,c}^{(r)}, \dots, h_{|\mathcal{V}|,c}^{(r)}]$ and
$H_s^{(r)} = [h_{1,s}^{(r)}, \dots, h_{|\mathcal{V}|,s}^{(r)}]$ denote the node feature matrices for relation $r$. These refined embeddings, $H_c$ and $H_s$, provide disentangled causal and shortcut representations for subsequent intervention.

\subsubsection{Disentanglement and Causal Intervention}
Until now, we have constructed the initial causal and shortcut graphs; however, these representations still need to be disentangled to separately capture causal and shortcut features from the input multimodal graph. Specifically, the causal graph is expected to encode task-relevant causal features. We apply a readout function followed by a predictor $\Phi_c$ to generate predictions, and define the supervised loss as the mean squared error (MSE) between predicted values and ground-truth labels:
\begin{equation}
\hat{y}_c = \Phi_c(\mathrm{Readout}(H_c)),
\label{eq:10}
\end{equation}
\begin{equation}
\mathcal{L}_{sup} = \frac{1}{N} \sum_{i=1}^{N} 
\left\| y_i - \hat{y}_{c,i} \right\|_2^2,
\label{eq:11}
\end{equation}
where \(N\) denotes the number of training samples, \(\hat{y}_{c,i}\) is the causal-branch prediction for the \(i\)-th sample, and \(y_i\) is the corresponding ground-truth label.


In contrast, the shortcut branch is designed to absorb unstable and non-causal information that is not necessary for robust sentiment prediction. Inspired by~\cite{SuCI2025,CAL2022}, we impose an auxiliary uniformity constraint on its prediction distribution, encouraging the shortcut branch to contain less task-discriminative semantic information. Specifically, we pass the shortcut representation through a separate predictor $\Phi_s$ and encourage its output distribution to be close to a uniform distribution. This is achieved by minimizing the Kullback–Leibler (KL) divergence:
\begin{equation}
\hat{p}_s = \mathrm{softmax}(\Phi_s(\mathrm{Readout}(H_s))),
\label{eq:12}
\end{equation}
\begin{equation}
\mathcal{L}_{unif} = \frac{1}{N} \sum_{i=1}^{N} 
\mathrm{KL}\left(y_{unif} \,\|\, \hat{p}_{s,i} \right),
\label{eq:13}
\end{equation}
where $y_{unif}$ denotes a uniform distribution over the discretized label space. If $C$ is the number of discretized classes from the original continuous labels, each entry of $y_{unif}$ is set to $1/C$.

Finally, to disentangle causal and shortcut features, we introduce a causal intervention loss based on backdoor adjustment~\cite{pearl2009causality}. As discussed in Sec.~\ref{A Causal Perspective on MSA}, the confounder $Z$ represents unobserved dataset-specific nuisance factors. Since $Z$ is neither directly observable nor exhaustively enumerable in real MSA datasets, we use the learned shortcut representations as empirical proxies for its effects. Accordingly, the shortcut memory bank $\hat{K}$ serves as an estimated stratification set of the latent confounder, where each stored representation corresponds to a potential shortcut context. To reduce computational overhead, we construct $\hat{K}$ online using shortcut representations from the current mini-batch. Specifically, following~\cite{CAL2022}, we sample shortcut representations $H_s^{(k)}$ from $\hat{K}$ and randomly combine them with the current causal representation. This simulates evaluating the same causal representation under different shortcut contexts, thereby approximating the expectation over confounding states in backdoor adjustment:
\begin{equation}
\hat{y}_{cs} = \Phi_{c}\left(\mathrm{Readout}(H_c) + \mathrm{Readout}(H_s^{(k)})\right),
\label{eq:14}
\end{equation}
\begin{equation}
\mathcal{L}_{intv} = \frac{1}{N \cdot |\hat{K}|}
\sum_{i=1}^{N} \sum_{k \in \hat{K}}
\left\| y_i - \hat{y}_{cs,i} \right\|_2^2,
\label{eq:15}
\end{equation}
where $H_s^{(k)}$ denotes the $k$-th shortcut representation sampled from the shortcut memory bank $\hat{K}$, and $\Phi_c$ is the prediction head shared with the causal branch. The final training objective is a weighted sum of the three losses:
\begin{equation}
\mathcal{L} = \mathcal{L}_{sup} 
+ \lambda \mathcal{L}_{unif} 
+ \beta \mathcal{L}_{intv},
\label{eq:16}
\end{equation}
where the hyperparameters $\lambda$ and $\beta$ control the weights of the shortcut uniformity loss and the causal intervention loss, respectively. As optimization proceeds, the causal branch is encouraged to learn stable causal features, while the shortcut branch is regularized to capture potential shortcut patterns.


\section{Experiments}
In this section, we conduct extensive experiments to answer the following research questions:  
\textbf{RQ1}: How does MMCI perform compared to existing state-of-the-art methods on standard benchmark datasets?  
\textbf{RQ2}: Can MMCI improve performance under out-of-distribution settings, and how does it compare with other causal-based approaches?  
\textbf{RQ3}: How does each component of MMCI affect its overall performance?

Due to space limitations, the details of the datasets, evaluation metrics, and feature extraction are provided in Appendix.

\begin{table}[t]
\centering
\renewcommand\tabcolsep{1.3pt}
\renewcommand\arraystretch{1.10}
\caption{Dataset statistics, data splits, and hyper-parameter settings for the three multimodal datasets.}
\label{tab:para}
\begin{tabular}{cccc}
\hline
Dataset & CMU-MOSI & CMU-MOSEI & CH-SIMS \\
\hline
\hline
Train & 1,284 & 16,326 & 1,368 \\
Valid & 229 & 1,871 &456 \\
Test  & 686 & 4,659 & 457 \\
\hline
\hline    
Batch Size & 8 & 32 & 16 \\
Epochs & 50 & 15 & 50 \\
Warm-up & \checkmark & \checkmark & \checkmark \\
Initial Learning Rate & $1\times10^{-5}$ & $1\times10^{-5}$ & $9\times10^{-5}$ \\
Optimizer & AdamW & AdamW & AdamW \\
Dropout Rate & 0.3 & 0.3 & 0.3 \\
Shared Feature Dimension $d$ & 256 & 256 & 256 \\
Shortcut Uniformity Loss Weight $\lambda$ & 0.2 & 0.4 & 0.2 \\
Causal Intervention Loss Weight $\beta$ & 0.6 & 0.2 & 0.6 \\
\hline
\end{tabular}
\end{table}





\subsection{Experimental Details}
We implement the proposed MMCI framework in PyTorch on an NVIDIA RTX A6000 GPU (48GB), with CUDA 11.6 and PyTorch 1.13.1. Training uses the AdamW optimizer \cite{AdamW}. Detailed hyper-parameter settings are listed in Table \ref{tab:para}. We conduct a grid search with 40 random iterations to select optimal settings. The batch size is chosen from $\{8, 16, 32, 64\}$, while the learning rate and fusion feature dimension are searched in $\{1\mathrm{e}{-5}, 2\mathrm{e}{-5}, 3\mathrm{e}{-5}, 6\mathrm{e}{-5}, 9\mathrm{e}{-5}\}$ and $\{64, 128, 256, 512\}$, respectively. The dropout rate is selected from $\{0.1, 0.2, 0.3, 0.4\}$, and $\lambda$ and $\beta$ are tuned within $\{0.1, 0.2, \ldots, 1.0\}$. Other hyperparameters are fixed. We adopt early stopping, terminating training if validation loss does not decrease for 20 consecutive epochs. The final model is selected based on the lowest MAE on the validation set.

\begin{table*}
\centering
\renewcommand\tabcolsep{5pt} 
\renewcommand\arraystretch{1.10}
\caption{Comparison on the CMU-MOSI and CMU-MOSEI datasets. $^\dag$ denotes results reported in the original papers, $^\S$ denotes results from \cite{DAFSA2026}, while the remaining results are obtained from our own experiments. The best results are highlighted in bold, and the second-best results are underlined.}
\label{tab:mosi_mosei}
\begin{tabular}{cccccccccccc}
\toprule
\multirow{2}{*}{Method} & \multicolumn{5}{c}{CMU-MOSI} & & \multicolumn{5}{c}{CMU-MOSEI} \\ 
\cmidrule{2-6} \cmidrule{8-12}
& Acc2($\uparrow$) & F1($\uparrow$) & Acc7($\uparrow$)  & MAE($\downarrow$) & Corr($\uparrow$) & & Acc2($\uparrow$) & F1($\uparrow$) & Acc7($\uparrow$) & MAE($\downarrow$) & Corr($\uparrow$) \\ 

\midrule

Graph-MFN$^\S$~\cite{CMU-MOSEI2018}
& 76.8 / 78.0 & 76.7 / 78.0 & 34.4 & 0.966 & 0.653 &
& 80.3 / 83.8 & 80.8 / 83.7 & 52.0 & 0.564 & 0.727 \\

MTAG$^\S$~\cite{MTAG2021}
& 80.8 / 83.1 & 80.4 / 82.8 & 25.9 & 0.879 & 0.706 &
& 81.5 / 84.6 & 81.7 / 84.5 & 52.6 & 0.559 & 0.735 \\

HGraph-CL$^\dag$~\cite{HGraph-CL2022} 
&84.3 / 86.2	&84.6 / 86.2 & -	&0.717 	&0.799    &  
&\underline{84.5} / 85.9 	&\underline{84.5} / 85.8 & -	&0.527 	&0.769 \\ 


ALMT~\cite{ALMT2023}
&82.4 / 84.5	&82.2 / 84.4 & 45.9	&0.741 	&0.776  &
&80.7 / 84.6	&81.3 / 84.7 & 52.7	&0.543 	&0.761  \\

TETFN$^\dag$~\cite{TETFN2023}
&84.1 / 86.1	&83.8 / 86.1 & -	&0.717 	&0.800  &
&84.3 / 85.2	&84.2 / 85.3 &-	&0.551 	&0.748  \\

DecAlign$^\dag$~\cite{DecAlign2026}
& - / 85.8 & - / 85.8 & 45.1 & 0.735 & 0.811 &
& - / 86.5 & - / 86.1 & \textbf{55.0} & 0.543 & 0.768 \\

ConFEDE~\cite{ConFEDE2023}
 &83.2 / 85.1	&83.2 / 85.2 & 43.3	&0.728 	&0.784  &
&81.2 / 85.7	&81.8 / 85.6 & 52.7	&0.538 	&0.772  \\  

DMD~\cite{DMD2023}
&82.2 / 84.3	&82.1 / 84.3 & 44.9	&0.726 	&0.788 &
&80.6 / 84.6	&81.1 / 84.6 & 52.8	&0.538	&0.768  \\ 

GLoMo~\cite{GLoMo2024}
&82.3 / 84.9	&82.1 / 84.8 & 45.6	&0.734 	&0.778  &
&82.4 / 85.6	&82.8 / 85.5 & 52.3	&0.558 	&0.748  \\

DEVA$^\dag$~\cite{DEVA2025}  
 & 84.4 / 86.3 & 84.5 / 86.3 & 46.3 & 0.730 & 0.787 & 
 & 83.3 / 86.1 & 82.9 / 86.2 & 52.3 & 0.541 & 0.769 \\  

DLF~\cite{DLF2025}
&82.2 / 84.8	&82.1 / 84.7 & \underline{46.7} &0.731	&0.787  &
&82.0 / 85.2&82.4 / 85.2  & 52.1 &0.543	&0.764  \\  

EMOE~\cite{EMOE2025}
&82.9 / 84.8	&82.9 / 84.8	 & 45.2	&0.723 	&0.790   &
&81.6 / 85.0	&82.1 / 85.0	 & 52.5	 &0.542	&0.760  \\ 

MLCL$^\dag$~\cite{MLCL2025}
&84.1 / 86.4	&83.9 / 86.3	 &46.6	&0.701 	&0.798  &
&84.1 / 86.3	&84.3 / 86.2	 &53.2	 &0.551	&0.756  \\ 

PSA-MF$^\dag$~\cite{PSA-MF2026}
&83.7 / 86.4	&83.3 / 86.2	 &46.5	&0.686 	&0.807 &
&83.8 / 86.3	&84.2 / 86.3	 &\textbf{55.0}	 &\textbf{0.521}	&0.774 \\ 

C-MIB~\cite{MIB2023}
 &	85.4 / 87.2 &	85.3 / 87.2 &  \textbf{47.6}	&\underline{0.650} 	&0.840 &
&83.7 / \underline{86.6}	&84.1 / \underline{86.6} &\underline{53.8}	&\underline{0.526} 	&0.779 \\

ITHP~\cite{ITHP2024}
& \underline{86.1} / \underline{88.2} & \underline{86.0} / \underline{88.2} & 46.3 & 0.654 & \underline{0.844} &  
& 82.3 / 86.2 & 82.9 / 86.3 & 51.6  & 0.556 & \underline{0.781} \\



\midrule
\textbf{MMCI (Ours)}
& \textbf{87.4} / \textbf{89.3} &  \textbf{87.4} / \textbf{89.3} & \textbf{47.6} & \textbf{0.616} & \textbf{0.856}   &
& \textbf{84.7} / \textbf{87.7} & \textbf{85.0} / \textbf{87.6} & 53.3 & \textbf{0.521} & \textbf{0.790} \\  

\bottomrule

\end{tabular}  
\end{table*}

\begin{table}[t]
\centering
\renewcommand\tabcolsep{2.5pt} 
\renewcommand\arraystretch{1.10}
\caption{Comparison on the CH-SIMS dataset. $^\dag$ indicates that the results are taken from~\cite{DEVA2025}, while the remaining results are reported in~\cite{MMSA2022}.}
\label{tab:sims}
\begin{tabular}{cccccccc}
\toprule
\multirow{2}{*}{Method} & \multicolumn{6}{c}{CH-SIMS} \\ 
\cmidrule{2-7}
& Acc-5($\uparrow$) & Acc-3($\uparrow$) & Acc-2($\uparrow$) & F1($\uparrow$) & MAE($\downarrow$) & Corr($\uparrow$) \\ 

\midrule

TFN~\cite{TFN2017}   & 39.3 & 65.1     & 78.4 & 78.6 & 0.432 & 0.591 \\
LMF~\cite{LMF2018}  &40.5  &  64.7     & 77.8 & 77.9 & 0.441 & 0.576 \\
MFN~\cite{MFN2018} &39.5  &  \underline{65.7}     & 77.9 & 77.9 & 0.435 & 0.582 \\
MulT~\cite{MulT2019}  &37.9  &  64.8    & 78.6 & 79.7 & 0.453 & 0.564 \\
Self-MM~\cite{Self-MM2021} & 41.5  &  65.5  & \underline{80.0} & \underline{80.4} & 0.425 & \underline{0.595} \\
CENet~\cite{CENet2022} & 33.9  &  62.6  & 77.9 & 77.5 & 0.471 & 0.540 \\


ALMT$^\dag$~\cite{ALMT2023}  & 40.7 &  65.0    & 78.6 & 78.9 & 0.450 & 0.535 \\
DEVA$^\dag$~\cite{DEVA2025}  & \underline{43.1} & 65.4   & 79.6 & 80.3 & \underline{0.424} & 0.583 \\
\midrule
\textbf{MMCI (Ours)} & \textbf{43.5} &  \textbf{66.3}   & \textbf{80.3} & \textbf{80.6} & \textbf{0.412} & \textbf{0.602}  \\

\bottomrule

\end{tabular}
\end{table}




\begin{table}[t]
\centering
\renewcommand\tabcolsep{4pt} 
\renewcommand\arraystretch{1.10}
\caption{Comparison on the OOD version of the CMU-MOSI dataset. Results marked with $^\dag$ are from the original papers, and $^\S$ indicates results from our experiments. Other results are taken from \cite{CLUE2022}.}
\label{tab:mosi_ood}
\begin{tabular}{cccccc}
\toprule
\multirow{2}{*}{Method} & \multicolumn{5}{c}{CMU-MOSI (OOD)} \\ 
\cmidrule{2-6}
& Acc7($\uparrow$) & Acc2*($\uparrow$) & Acc2($\uparrow$) & F1*($\uparrow$) & F1($\uparrow$) \\ 
 
\midrule
MulT~\cite{MulT2019}   & 29.8 & 75.0 &76.7 & 74.8 &76.5 \\
MAG-BERT~\cite{MAG2020} & 39.9 & 75.6 &77.3 & 75.5 &77.3 \\
MISA~\cite{MISA2020}  & 38.0 & 75.9 &77.4 & 75.8 &77.4 \\
Self-MM~\cite{Self-MM2021}   & 40.2 & 76.7 &78.1 & 76.7 &78.1 \\
CLUE~\cite{CLUE2022}   & 41.8 & 78.8 &79.9 & 78.8 &79.9    \\
GEAR$^\dag$~\cite{GEAR2023}  & -    & \underline{80.5} &\underline{82.1} & \underline{80.4} &\underline{82.1}    \\
MulDeF$^\dag$~\cite{MulDeF2024}  &  \underline{42.9} & 79.8 &81.4 & 79.9 &81.5 \\ 
ITHP$^\S$~\cite{ITHP2024} &41.3 & 79.5 & 81.3 & 79.5 &81.3 \\ 
\midrule

\textbf{MMCI (Ours)} & \textbf{44.5} & \textbf{81.2} &\textbf{83.3} & \textbf{81.2} &\textbf{83.3} \\

\bottomrule

\end{tabular}
\end{table}

\subsection{Performance on Standard MSA Datasets (RQ1)}
We compare MMCI with current state-of-the-art MSA methods on the CMU-MOSI and CMU-MOSEI datasets, including Graph-MFN~\cite{CMU-MOSEI2018}, MTAG~\cite{MTAG2021}, HGraph-CL~\cite{HGraph-CL2022}, ALMT~\cite{ALMT2023}, TETFN~\cite{TETFN2023}, DecAlign~\cite{DecAlign2026}, ConFEDE~\cite{ConFEDE2023}, DMD~\cite{DMD2023}, GLoMo~\cite{GLoMo2024}, DEVA~\cite{DEVA2025}, DLF~\cite{DLF2025}, EMOE~\cite{EMOE2025}, MLCL~\cite{MLCL2025}, PSA-MF~\cite{PSA-MF2026}, C-MIB~\cite{MIB2023} and ITHP~\cite{ITHP2024}. The details of these methods are provided in the Appendix.

Table~\ref{tab:mosi_mosei} summarizes the performance comparison between the proposed MMCI and various competitive baselines. The results demonstrate that MMCI consistently achieves state-of-the-art performance across almost all evaluation metrics on both CMU-MOSI and CMU-MOSEI. Specifically, on the CMU-MOSI dataset, MMCI notably surpasses the information bottleneck-based models ITHP~\cite{ITHP2024} and C-MIB~\cite{MIB2023}, outperforming ITHP by 1.3\% in Acc7 and achieving improvements of up to 2.1\% in Acc2 and F1 over C-MIB. MMCI also demonstrates superior performance over HGraph-CL~\cite{HGraph-CL2022}, which relies on contrastive learning between unimodal and multimodal graph representations, while our causal debiasing approach captures more robust sentiment features. Furthermore, the simultaneous reduction in MAE and enhancement in Acc2 (including neutral samples) underscores MMCI's precision in fine-grained sentiment estimation. This trend extends to the larger CMU-MOSEI dataset; MMCI still demonstrates clear advantages, achieving improvements of 1.5\%, 1.2\%, and 1.3\% over ITHP in Acc7, Acc2, and F1, respectively, while maintaining strong performance against other baselines across most metrics.

We further evaluate MMCI on the Chinese multimodal benchmark CH-SIMS dataset to assess its cross-lingual generalization. As shown in Table~\ref{tab:sims}, MMCI consistently outperforms all baselines across all evaluation metrics, including the recent state-of-the-art model DEVA~\cite{DEVA2025}. While DEVA leverages textual sentiment descriptions to guide the progressive fusion of audio and visual modalities, it still relies on feature-level alignment. In contrast, MMCI models causal and shortcut relationships within multi-relational multimodal graphs, helping reduce reliance on spurious correlations and capture more robust sentiment information. These results indicate the superior generalizability of MMCI in multilingual scenarios.

\subsection{Performance on the OOD Testing (RQ2)}
Table~\ref{tab:mosi_ood} presents the performance comparison between MMCI and other methods under the OOD testing setting. From this table, we observe the following: \textbf{i)} All methods perform worse under the OOD setting than on standard datasets, indicating that spurious correlations undermine generalization. \textbf{ii)} Under the OOD setting, MMCI significantly outperforms baseline methods based on conventional multimodal fusion techniques and further enlarges its advantage over ITHP. Specifically, the improvements in Acc2 increase from (1.3\%, 1.1\%) to (1.7\%, 2.1\%), and in F1 from (1.4\%, 1.1\%) to (1.7\%, 2.0\%), demonstrating that our causal debiasing approach achieves superior generalization. \textbf{iii)} Compared with causal-based baselines such as CLUE~\cite{CLUE2022}, GEAR~\cite{GEAR2023}, and MulDeF~\cite{MulDeF2024}, our model maintains outstanding performance and consistently outperforms all competitors across all metrics. While these methods mitigate spurious correlations through explicit causal intervention, counterfactual reasoning, or bias reweighting strategies, MMCI focuses on addressing spurious correlations both within and across modalities. This enables more effective suppression of diverse biases and leads to stronger OOD generalization.

\begin{figure}[t]
    \centering
    \includegraphics[width=\linewidth]{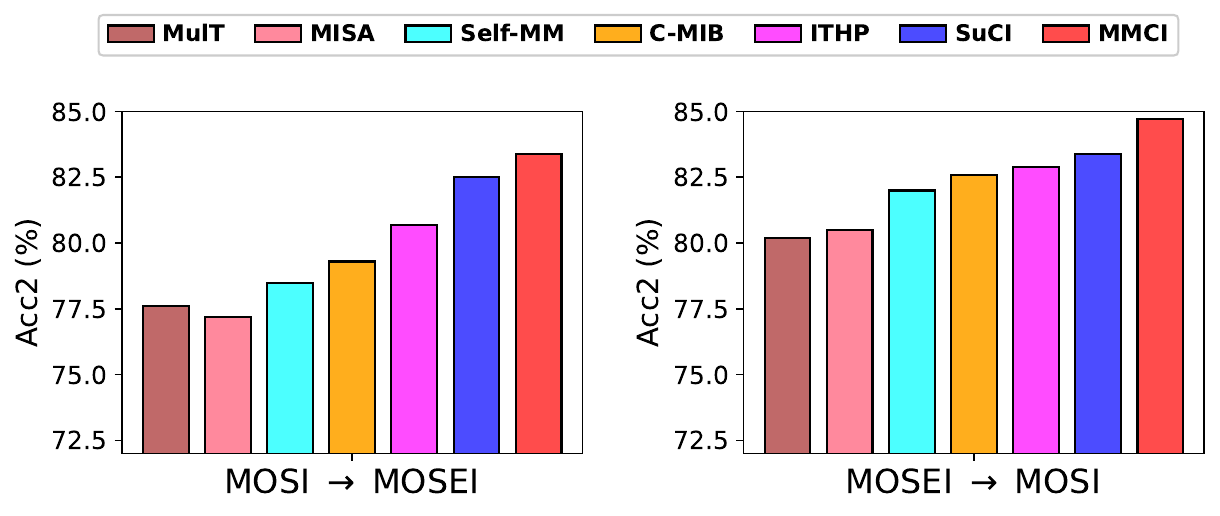}
    \caption{Cross-dataset evaluation of methods trained on the CMU-MOSI and CMU-MOSEI training sets and tested on the corresponding opposite datasets, including both MOSI$\rightarrow$MOSEI and MOSEI$\rightarrow$MOSI transfer settings. The results of ITHP and C-MIB are obtained from our own implementations, while the remaining results are adopted from~\cite{SuCI2025}.}
    \label{fig:cross}
\end{figure}

\subsection{Cross-dataset Evaluation (RQ2)}
Since CMU-MOSI and CMU-MOSEI follow the same annotation protocol, we conduct cross-dataset evaluation following Xu et al.~\cite{SuCI2025}. As shown in Fig.~\ref{fig:cross}, models trained on CMU-MOSI are directly evaluated on CMU-MOSEI, and vice versa. This setting provides a naturally occurring OOD scenario without introducing artificial distribution shifts, enabling a more realistic evaluation of model generalization and the impact of dataset-specific confounding factors. Most methods exhibit noticeable performance degradation under this challenging setting, which is expected as distribution shifts may amplify spurious correlations learned from the training data. Nevertheless, MMCI achieves the best cross-dataset performance, outperforming the recent causal method SuCI~\cite{SuCI2025} by 0.9 and 1.3 percentage points, respectively. These consistent improvements suggest that MMCI mitigates dataset-specific biases and reduces reliance on dataset-specific shortcuts, thereby improving generalization under OOD shifts.

\subsection{Ablation Studies (RQ3)}
We perform systematic ablation studies on the CMU-MOSI dataset to evaluate the different designs and mechanisms incorporated in the proposed MMCI.

\subsubsection{The Importance of Modeling Intra- and Inter-modal Relations} To investigate their respective effects, we remove intra-modal relations by using a shared GAT for all three modalities, resulting in the ``w/o Intra-modal Relation'' setting. Similarly, we remove inter-modal relations to obtain the ``w/o Inter-modal Relation'' setting. As presented in Table~\ref{tab:4}, removing intra-modal relations leads to performance drops of 1.9\%, 3.8\%, and 3.8\% in Acc7, Acc2, and F1 scores, respectively. Removing inter-modal relations also results in declines of 1.0\%, 0.6\%, and 0.6\% on the same metrics. These results demonstrate that both intra- and inter-modal relations help capture biases under different types of relations, with intra-modal biases having a greater impact.

\subsubsection{The Importance of Causal and Shortcut Attention}
To separately examine the contributions of causal and shortcut attention, we replace the corresponding learned edge-attention scores with a neutral value of 0.5 while keeping the graph structure and other components unchanged, resulting in the ``w/o Causal Attention'' and ``w/o Shortcut Attention'' variants. As shown in Table~\ref{tab:4}, removing causal attention leads to consistent performance degradation, indicating its role in capturing stable, task-relevant relations. In contrast, removing shortcut attention mainly affects performance on polarity-sensitive metrics (e.g., Acc2), suggesting that it helps identify and mitigate spurious relations during learning. When both attention branches are removed, the performance drops further, representing the most significant degradation among all variants. These results demonstrate that causal and shortcut attention play distinct yet complementary roles, and that their joint modeling is crucial for effective feature disentanglement and causal intervention.

\begin{figure*}[t]
    \centering
    \subfloat[Effect of $\lambda$ and $\beta$ on the CMU-MOSI dataset.]{
        \begin{minipage}[b]{0.48\linewidth}
            \centering
            \includegraphics[width=0.48\linewidth]{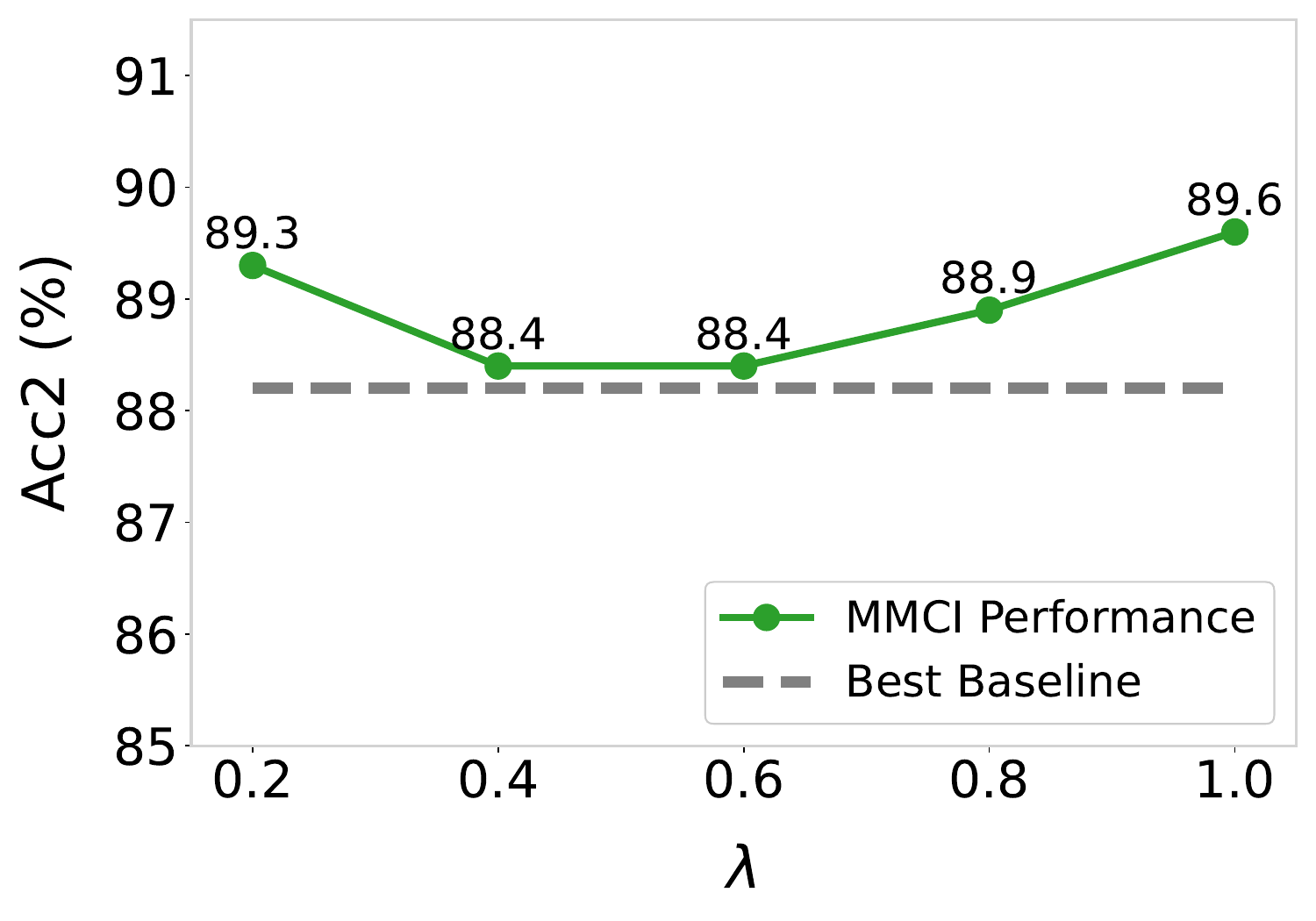}
            \hfill
            \includegraphics[width=0.48\linewidth]{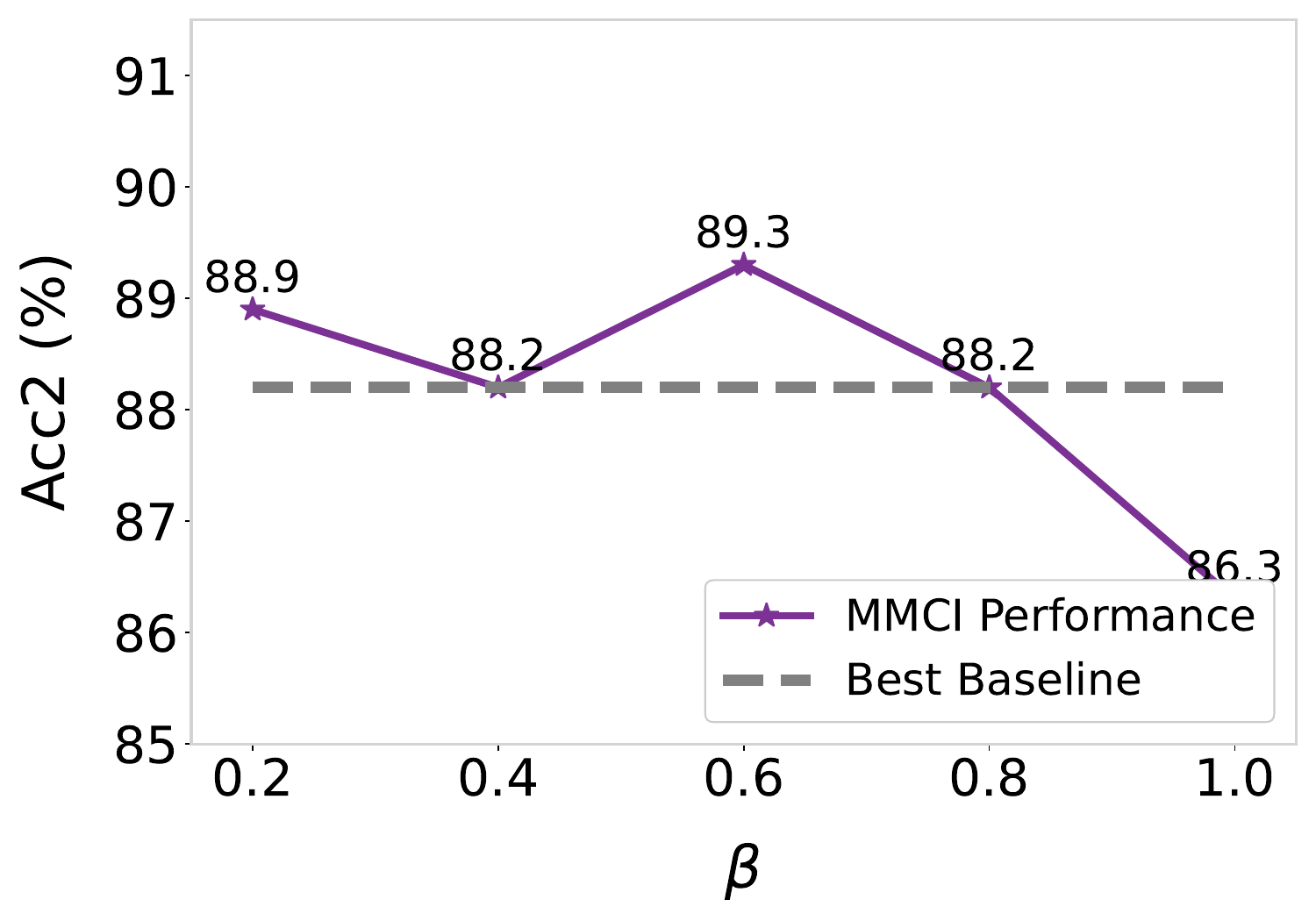}
        \end{minipage}
        \label{fig:para_mosi}
    }
    \hfill
    \subfloat[Effect of $\lambda$ and $\beta$ on the CMU-MOSEI dataset.]{
        \begin{minipage}[b]{0.48\linewidth}
            \centering
            \includegraphics[width=0.48\linewidth]{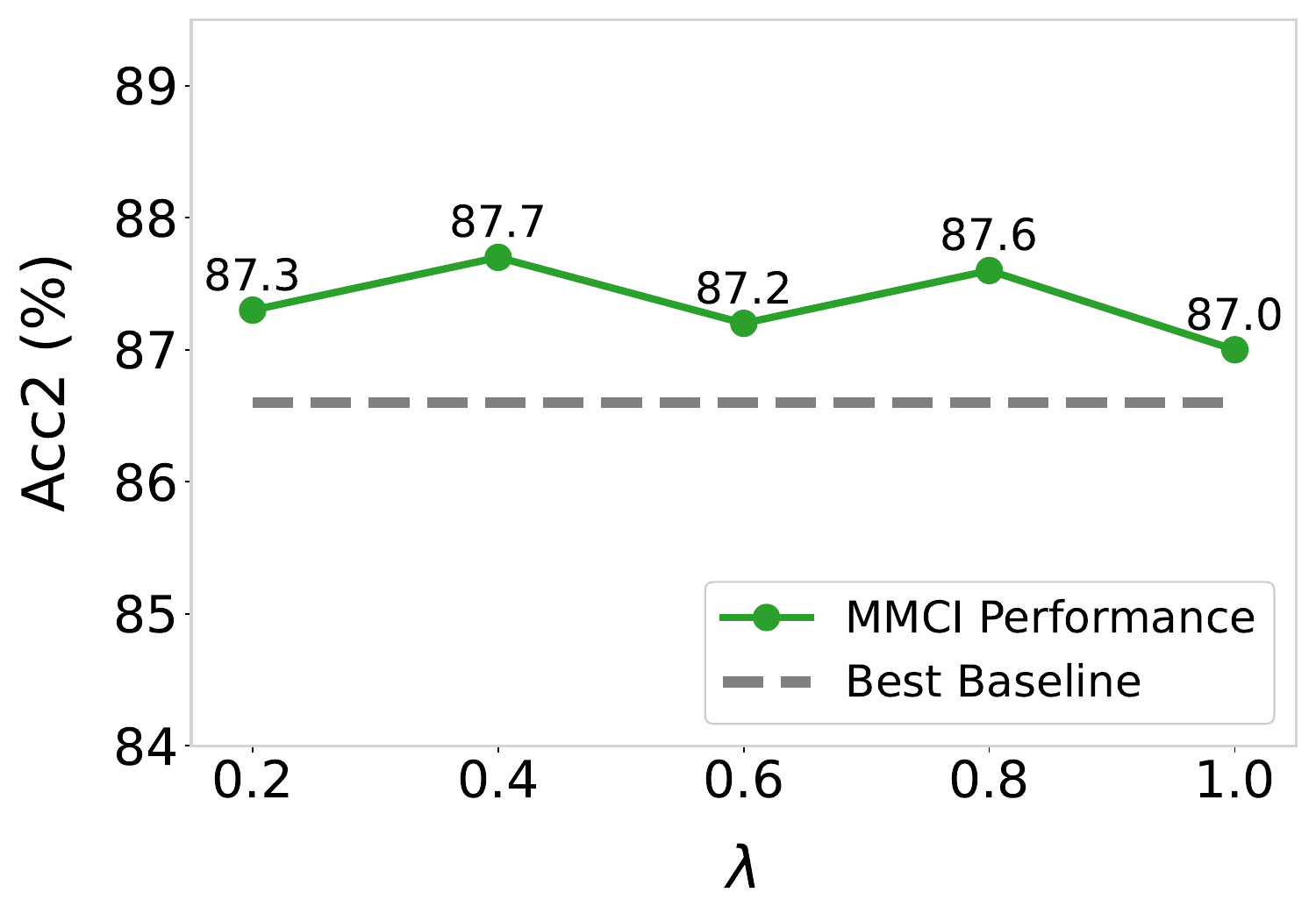}
            \hfill
            \includegraphics[width=0.48\linewidth]{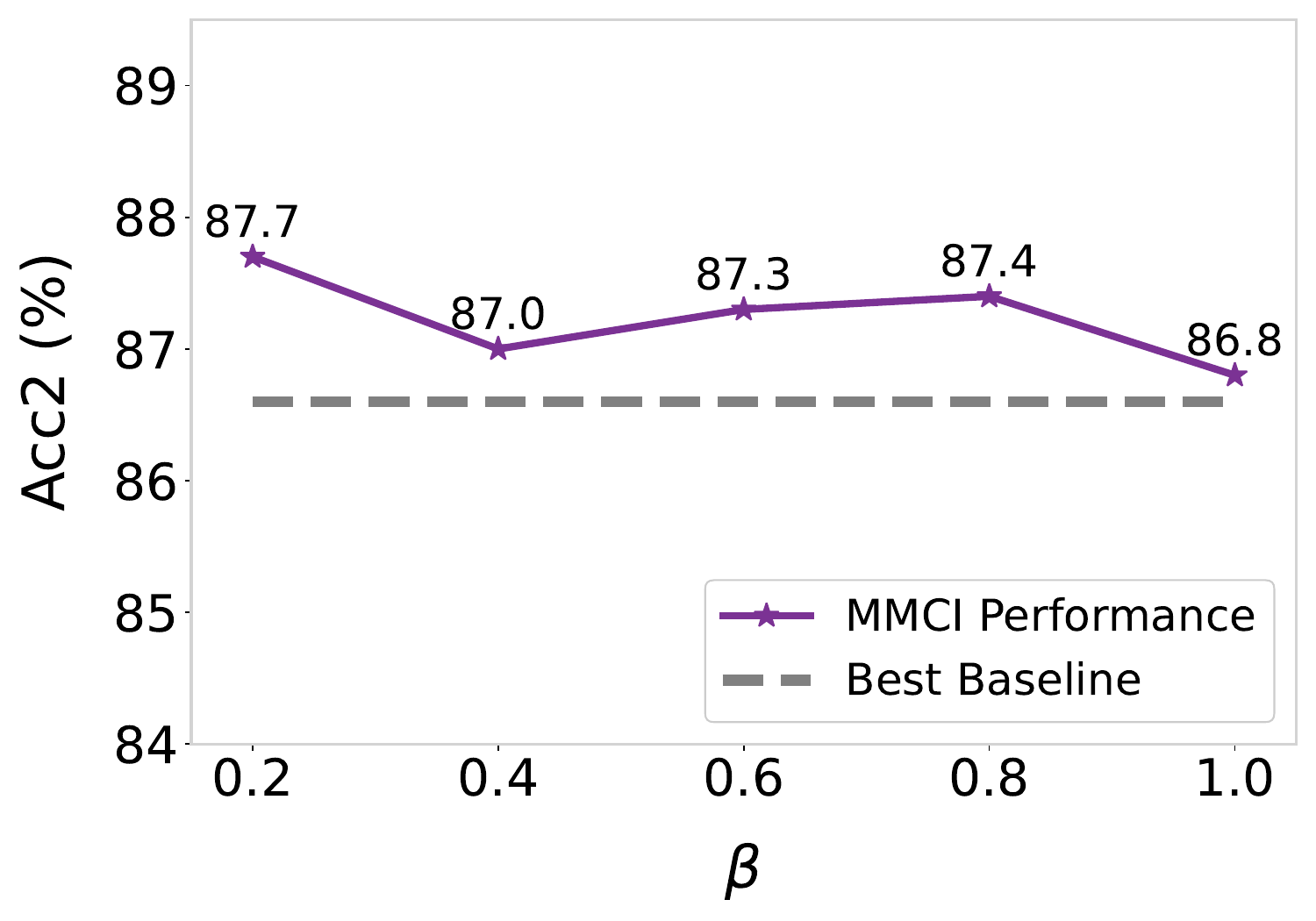}
        \end{minipage}
        \label{fig:para_mosei}
    }

    \caption{Parameter sensitivity analysis of disentanglement weight $\lambda$ and causal intervention weight $\beta$. The gray dashed line denotes the best baseline.}
    \label{fig:para}
\end{figure*}

\begin{figure}[t]
    \centering
    \includegraphics[width=0.49\linewidth]{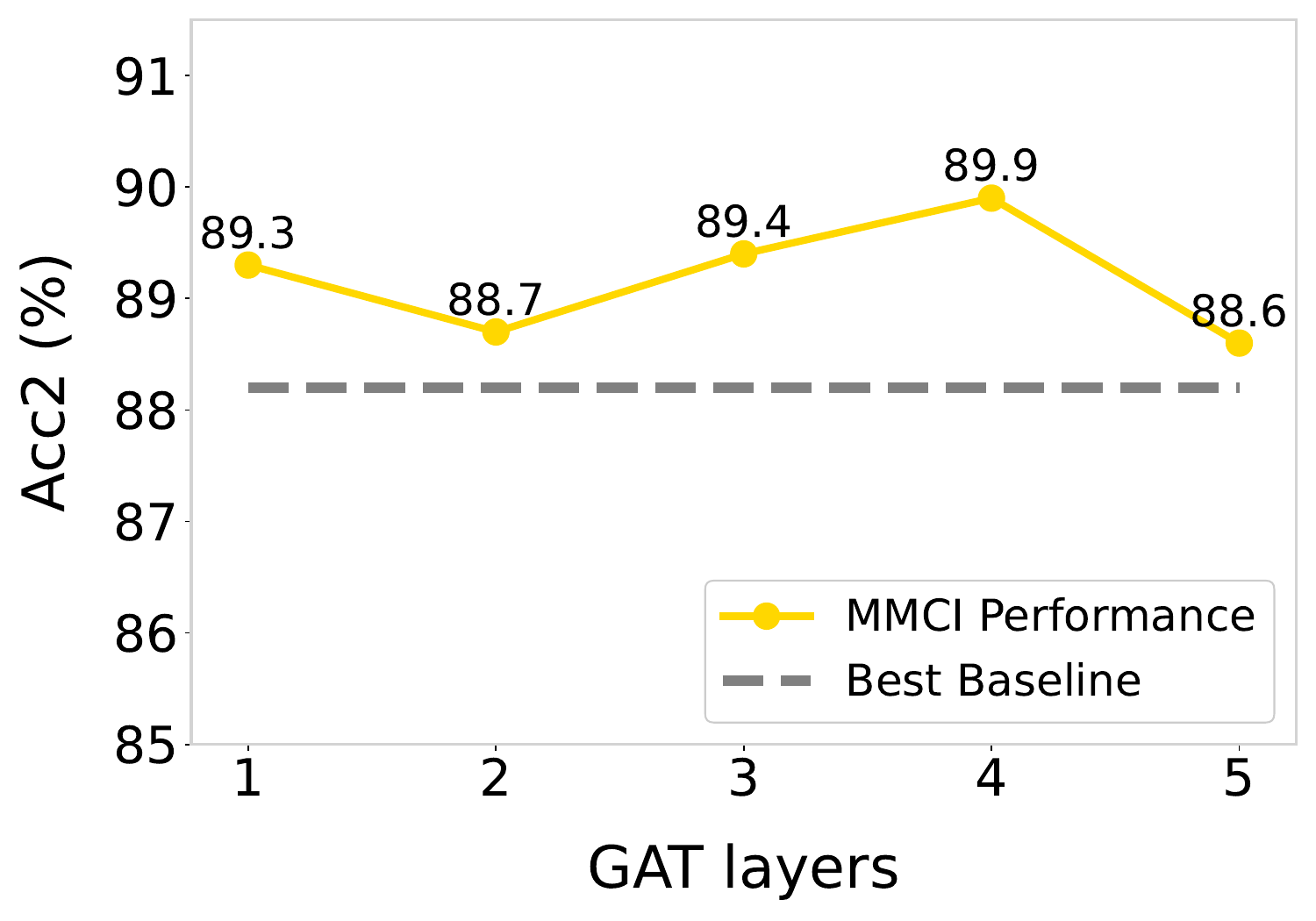}
    \hfill
    \includegraphics[width=0.49\linewidth]{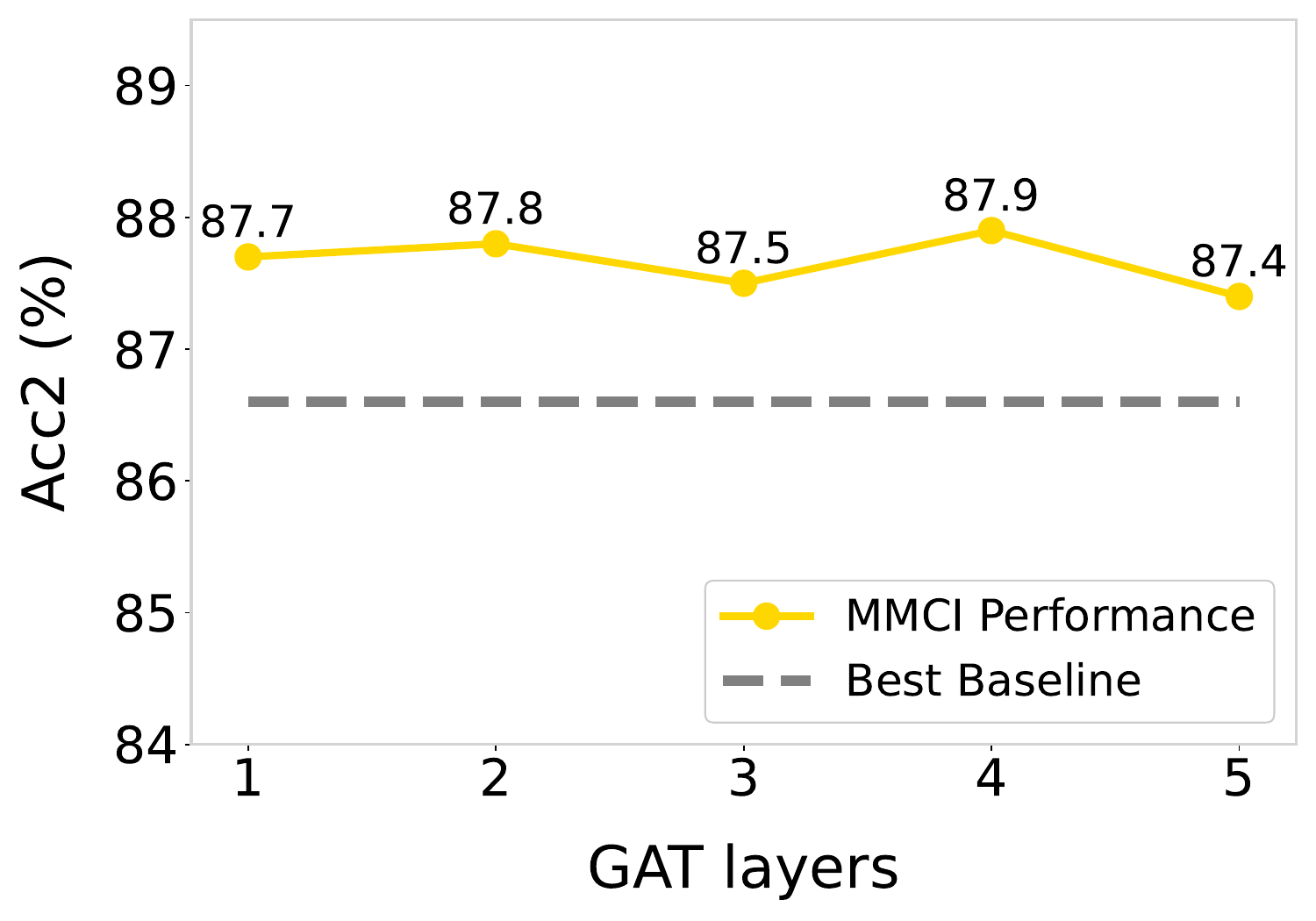}
    \caption{Effect of graph attention layers on CMU-MOSI (left) and CMU-MOSEI (right).}
    \label{fig:gat_layers}
\end{figure}

\subsubsection{The Importance of Disentanglement}
In the ``w/o Disentanglement'' setting, we remove the disentanglement process and evaluate the model without explicitly separating causal and shortcut features. Results show that the model's performance drops by 0.3\%, 0.9\%, and 0.9\% on Acc7, Acc2, and F1, respectively, validating the effectiveness of the proposed disentanglement strategy. Although the degradation appears modest, this is mainly because the initial shortcut graph exhibits limited mutual information with labels, restricting the extent to which disentanglement can remove semantically relevant information. This is further supported by the ``w/o Shortcut Supervision'' setting, where the shortcut branch is trained using a standard MSE loss instead of the proposed supervision strategy, reducing MMCI to a conventional multi-relational multimodal model without explicit shortcut guidance. In this case, performance drops significantly by 7.3\%, 6.1\%, and 6.0\% on Acc7, Acc2, and F1, respectively, clearly demonstrating the importance of disentangling causal and shortcut features for mitigating spurious correlations.


        




\begin{table}[t]
\centering
\caption{Ablation experiments on the CMU-MOSI dataset.}
\renewcommand\tabcolsep{5pt} 
\renewcommand\arraystretch{1.10}
\label{tab:4}
\begin{tabular}{lccc}
\toprule
Method
 & Acc2($\uparrow$) & F1($\uparrow$) & Acc7($\uparrow$) \\ 

\midrule
        
w/o Intra-modal Relation & 83.6 / 85.5 & 83.6 / 85.5  & 45.7 \\
w/o Inter-modal Relation   & 87.2 / 88.7 & 87.1 / 88.7 & 46.6 \\ 
\midrule
w/o Causal Attention & 86.3 / 87.8 & 86.3 / 87.8  & 45.8 \\
w/o Shortcut Attention & 86.9 / 88.4 & 86.8 / 88.4  & \textbf{48.0} \\
w/o Dual Attention Assignment & 84.9 /86.1 & 85.0 / 86.2  & 45.4 \\
\midrule
w/o Disentanglement  & 86.6 / 88.4 & 86.5 / 88.4  & 47.3\\
w/o Shortcut Supervision & 81.9 / 83.2 & 81.9 / 83.3&   40.3  \\
\midrule
w/o Causal Intervention & 85.1 / 86.9 & 85.1 / 86.9   & 42.8\\ 
w/o Random Combination & 86.6 / 87.9 & 86.6 / 87.9   & 46.0\\ 
\midrule

\textbf{MMCI (Ours)} 
 & \textbf{87.4} / \textbf{89.3} & \textbf{87.4} / \textbf{89.3} & 47.6\\  

\bottomrule
\end{tabular}

\end{table}

\subsubsection{The Importance of Causal Intervention}
In the ``w/o Causal Intervention'' setting, we remove the stratification and recombination of shortcut features, leading to consistent performance degradation and confirming the necessity of intervening on shortcut features.
We further introduce the ``w/o Random Combination'' variant, which retains causal and shortcut representations but replaces random cross-sample combination with a deterministic sample-wise pairing strategy. In this setting, each causal representation is exposed only to its corresponding shortcut context, rather than diverse shortcut patterns from the estimated confounder distribution. Compared with the full MMCI, this variant shows weaker performance, indicating that random combination helps approximate shortcut stratification and promotes robustness across different confounding contexts. Nevertheless, ``w/o Random Combination'' still outperforms ``w/o Causal Intervention'', showing that even deterministic interaction provides useful intervention signals, while random mixing further enhances robustness by exposing causal features to diverse shortcut variations.

\subsection{Parameter Sensitivity Analysis}
According to Eq.~\eqref{eq:16}, $\lambda$ controls the strength of disentanglement between causal and shortcut features, while $\beta$ governs the intensity of causal intervention. To investigate their effects, we fix one coefficient at its default value and vary the other within the range $[0.2, 1.0]$ with a step size of 0.2. Experiments are conducted on both the CMU-MOSI and CMU-MOSEI datasets, and the results are presented in Fig.~\ref{fig:para}. Overall, the model is relatively insensitive to variations in $\lambda$, but more sensitive to $\beta$. When $\beta$ is set to a relatively small value (e.g., 0.2), the model achieves strong performance. However, as $\beta$ increases (e.g., to 1.0), performance drops significantly on both datasets. This is because overly strong causal intervention overemphasizes the removal of confounding factors, thereby disrupting the feature fusion process essential for effective multimodal understanding. Interestingly, setting $\lambda$ to 1.0 yields performance that even surpasses the reported results, suggesting that careful hyperparameter tuning can further improve MMCI. In general, our method is not overly sensitive to these hyperparameters. As shown in Fig.~\ref{fig:para}, MMCI consistently outperforms the strongest baseline across a wide range of hyperparameter settings.

Beyond the loss coefficients, we further examine the sensitivity of MMCI to the number of graph attention layers by varying the depth from 1 to 5 while keeping other settings unchanged. As shown in Fig.~\ref{fig:gat_layers}, performance remains relatively stable across different depths. Although one layer is used as the default lightweight configuration, deeper graph propagation can further improve performance. In particular, the four-layer configuration achieves the best results on both datasets, improving Acc2 by 0.6 and 0.2 percentage points over the default setting on CMU-MOSI and CMU-MOSEI, respectively. However, the improvement is not monotonic, and five layers lead to performance degradation, possibly due to redundant information or over-smoothing caused by excessive message passing. The performance gains achieved by four layers are relatively modest, whereas stacking additional layers requires repeated relation-specific message passing and introduces substantially higher computational cost. Considering this performance--efficiency trade-off, we adopt one layer as the default configuration, while three or four layers can be considered when computational resources permit.


\begin{table}[t]
\centering
\renewcommand\tabcolsep{2.5pt} 
\renewcommand\arraystretch{1.10}
\caption{Acc2 performance under Gaussian and Salt-Pepper noise on the CMU-MOSI and CMU-MOSEI datasets.}
\label{tab:noise}
\begin{tabular}{cccccccc}
\toprule
\multirow{2}{*}{Method} & \multicolumn{3}{c}{Gaussian Noise} & & \multicolumn{3}{c}{Salt-Pepper Noise} \\
\cmidrule{2-4} \cmidrule{6-8}
 & $\epsilon=1$ & $\epsilon=5$ & $\epsilon=10$ & & $\epsilon=1$ & $\epsilon=5$ & $\epsilon=10$ \\
\midrule \midrule
\multicolumn{8}{c}{Noisy CMU-MOSI} \\ 
\midrule
C-MIB~\cite{MIB2023} & 84.6 & 78.5 & 68.4 &  &84.6 & 78.6 &67.3 \\
ITHP~\cite{ITHP2024} & 86.1 & 79.2 & 64.9 & & 82.8 &76.3 & 63.1 \\
QMF~\cite{QMF2023} & 84.1 & 76.3 & \textbf{69.5} & & 83.1 & \textbf{80.0} & 63.7 \\
\textbf{MMCI (Ours)} & \textbf{87.0} & \textbf{81.2} & 68.2 & & \textbf{86.3} & 73.7 & \textbf{67.5} \\
\midrule \midrule
\multicolumn{8}{c}{Noisy CMU-MOSEI} \\ 
\midrule
C-MIB~\cite{MIB2023} & 84.7 & 81.4 & 73.5 & & 85.9 & 83.3 & 73.0 \\
ITHP~\cite{ITHP2024} & 85.5 & 82.8 & 74.7 & & 85.6 & 80.4 & \textbf{75.5} \\
QMF~\cite{QMF2023} & 86.1 & 82.9 & 74.0 & & 85.9 & 83.1 & 74.2 \\
\textbf{MMCI (Ours)} & \textbf{86.7} & \textbf{83.2} & \textbf{75.2} & &\textbf{86.7} & \textbf{83.8} & 74.1 \\
\bottomrule
\end{tabular}
\end{table}

\subsection{Discussion on Noisy Modalities}

To simulate practical data-quality degradation caused by sensor errors, transmission artifacts, and imperfect preprocessing, we introduce controlled perturbations into modality representations to evaluate the robustness of MMCI. Specifically, Gaussian noise and Salt-Pepper noise are injected at three intensity levels, i.e., $\epsilon\in\{1,5,10\}$~\cite{QMF2023}. Gaussian noise models continuous random perturbations, while Salt-Pepper noise represents sparse impulsive corruption. We compare MMCI with information bottleneck-based methods C-MIB~\cite{MIB2023} and ITHP~\cite{ITHP2024}, as well as the uncertainty-aware fusion method QMF~\cite{QMF2023}. As shown in Table~\ref{tab:noise}, MMCI achieves the best performance in 9 out of 12 settings. On CMU-MOSI, it performs best under Gaussian noise with $\epsilon=1$ and $5$, and Salt-Pepper noise with $\epsilon=1$ and $10$. On CMU-MOSEI, MMCI achieves the best results under all Gaussian noise levels and Salt-Pepper noise with $\epsilon=1$ and $5$. These results demonstrate that MMCI is robust under most corruption scenarios. This may be attributed to causal--shortcut disentanglement, which reduces reliance on unstable features, and shortcut-based intervention, which encourages consistent predictions across corrupted contexts.

\begin{table}[t]
\centering
\renewcommand\tabcolsep{8pt} 
\renewcommand\arraystretch{1.10}
\caption{Performance comparison on the CMU-MOSI dataset. Methods based on BERT and DeBERTa are marked with subscripts ``b'' and ``d'', respectively. $^\dag$ indicates results obtained from our experiments, while other results are taken from \cite{ITHP2024}.}
\label{tab:deberta}
\begin{tabular}{lcccc}
\toprule
Method &  
 Acc2($\uparrow$) & F1($\uparrow$) & MAE($\downarrow$) & Corr($\uparrow$) \\ 

\midrule

Self-MM\textsubscript{b} \cite{Self-MM2021} 
& 84.0 & 84.4 & 0.713 & 0.798  \\

MMIM\textsubscript{b} \cite{MMIM2021} 
& 84.1 & 84.0 & 0.700 & 0.800  \\ 

MAG\textsubscript{b} \cite{MAG2020} 
& 84.2 & 84.1 & 0.712 & 0.796 \\

C-MIB$^\dag$\textsubscript{b} \cite{MIB2023} 
& 84.7	& 84.7	& 0.717 	& 0.795 		 \\

\midrule

Self-MM\textsubscript{d} \cite{Self-MM2021} 
& 55.1 & 53.5 & 1.440 & 0.158\\

MMIM\textsubscript{d} \cite{MMIM2021} 
& 85.8 & 85.9 & 0.649 & 0.829  \\ 

MAG\textsubscript{d} \cite{MAG2020} 
& 86.1 & 86.0 & 0.690 & 0.831 \\

C-MIB$^\dag$\textsubscript{d} \cite{MIB2023} 
& 87.2 & 87.2 & 0.650 & 0.840\\

ITHP$^\dag$\textsubscript{d} \cite{ITHP2024} 
& 88.2 & 88.2 & 0.654 & 0.844  \\

\midrule
\textbf{MMCI (Ours) }
& \textbf{89.3} & \textbf{89.3} & \textbf{0.616} & \textbf{0.856} \\   
\bottomrule
\end{tabular}%

\end{table}

\subsection{Discussion on the Pre-trained Language Model}
For our main task of MSA, following the state-of-the-art method ITHP \cite{ITHP2024}, we adopt DeBERTa~\cite{DeBERTa2020} as the pre-trained language model. In this section, we evaluate and analyze the impact of different pre-trained language models on overall performance. As shown in Table~\ref{tab:deberta}, models equipped with DeBERTa generally outperform their BERT-based counterparts. For example, the DeBERTa-based variant of MMIM~\cite{MMIM2021} achieves higher Corr (0.829 vs.\ 0.800) and lower MAE (0.649 vs.\ 0.700). Nevertheless, even with a stronger text encoder, these models still fall short of the performance achieved by our proposed MMCI. This suggests that simply adopting a more powerful text encoder is insufficient; instead, the design of effective debiasing mechanisms plays a crucial role in achieving state-of-the-art performance.

\subsection{Discussion on Unimodal and Bimodal Models}
In this subsection, we analyze the performance of unimodal and bimodal variants of MMCI. Prior studies have shown that the text modality is most informative for sentiment analysis, while audio and visual modalities provide complementary cues \cite{ConFEDE2023,MGT2025}. Accordingly, we focus on configurations involving text-only and text-based multimodal combinations. Table~\ref{tab:modality_combinations} summarizes the results, from which we draw the following observations:
Firstly, MMCI consistently outperforms baseline ITHP~\cite{ITHP2024} in both unimodal and bimodal settings, demonstrating the effectiveness and robustness of our debiasing strategy across modality combinations.
Secondly, bimodal models generally outperform unimodal ones, while the trimodal setting achieves the best overall performance. This confirms the advantage of integrating complementary information from multiple modalities.
Furthermore, the consistent improvements across modality settings indicate that MMCI remains effective even when certain modalities are unavailable, highlighting its robustness across different modality configurations.

\begin{table}[t]
\centering
\caption{Performance Comparison of MMCI and ITHP across various modality combinations on the CMU-MOSI dataset.}
\label{tab:modality_combinations}
\setlength\tabcolsep{5.6pt}
\renewcommand\arraystretch{1.1}
\begin{tabular}{cccccccc}
\toprule
\multicolumn{3}{c}{Modality} & \multicolumn{2}{c}{ITHP~\cite{ITHP2024}} & \multicolumn{2}{c}{\textbf{MMCI (Ours)}} \\
\cmidrule(lr){1-3} \cmidrule(lr){4-5} \cmidrule(lr){6-7}
\textit{T} & \textit{V} & \textit{A} 
& Acc7 ($\uparrow$) & Acc2 ($\uparrow$) & Acc7 ($\uparrow$) & Acc2 ($\uparrow$) \\
\midrule

$\checkmark$ & $\text{\sffamily x}$  & $\text{\sffamily x}$
& 42.3 & 85.3 / 87.0 & \textbf{46.0} & \textbf{86.0} / \textbf{87.9} \\  

$\checkmark$ & $\checkmark$ & $\text{\sffamily x}$
&43.5 & 85.4 / 87.5 & \textbf{48.0}  &  \textbf{86.9} / \textbf{88.5} \\ 

$\checkmark$ & $\text{\sffamily x}$  & $\checkmark$ 
&  46.7  &  84.8 / 86.7  & \textbf{47.4}  & \textbf{86.3} / \textbf{88.2} \\  

$\checkmark$ & $\checkmark$ & $\checkmark$
&46.3 & 86.1 / 88.2 &  \textbf{47.6} &  \textbf{87.4} / \textbf{89.3} \\ 
\bottomrule
\end{tabular}
\end{table}

\subsection{Visualization of the Features}
\begin{figure}[t]
    \centering
    \includegraphics[width=0.48\linewidth]{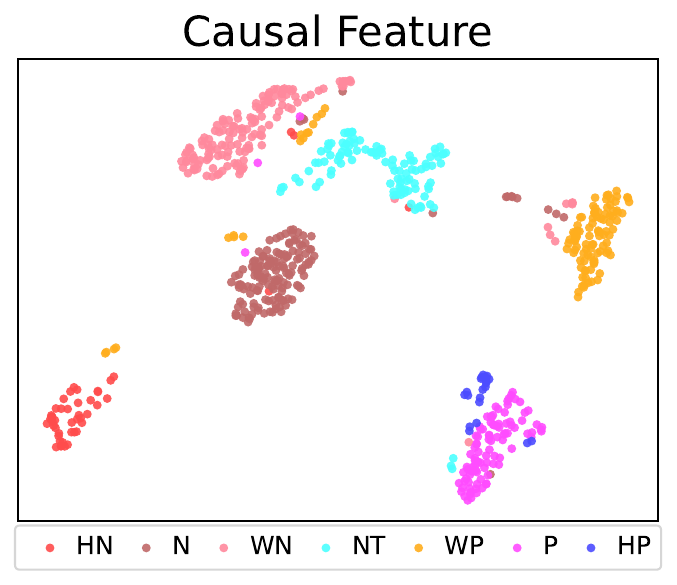}
    \hfill
    \includegraphics[width=0.48\linewidth]{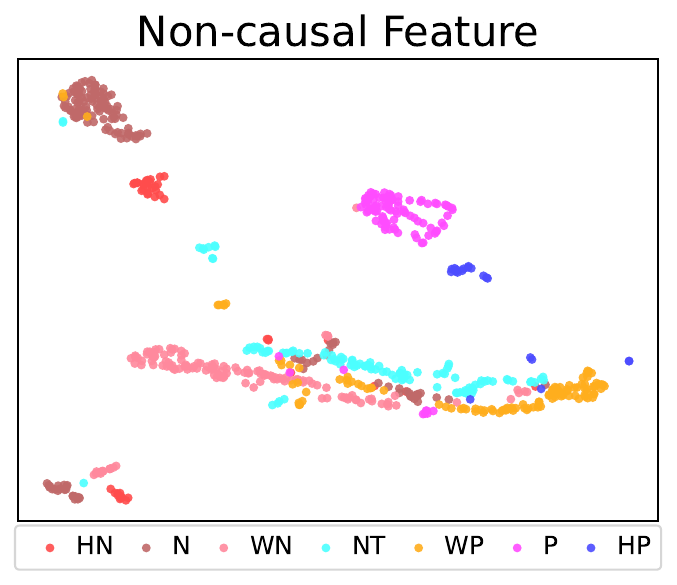}
    \caption{Visualization of causal features (left) and non-causal features (right). HN: Highly Negative; N: Negative; WN: Weakly Negative; NT: Neutral; WP: Weak Positive; P: Positive; HP: Highly Positive.}
    \label{fig:umap}
\end{figure}

To further examine the effectiveness of feature disentanglement, we visualize the causal and non-causal representations learned on CMU-MOSI. As shown in Fig.~\ref{fig:umap}, the causal features form compact and well-separated clusters across sentiment categories, indicating that the causal branch captures discriminative and sentiment-relevant semantics. In contrast, non-causal features are more intertwined and show limited category separability, particularly for sentiment states near neutral, where adjacent categories are difficult to distinguish. This observation supports the effectiveness of the uniformity constraint and causal intervention in reducing directly decodable label information in the non-causal branch while preserving invariant sentiment semantics in the causal branch. However, some non-causal features remain partially distinguishable under strong sentiment polarity, suggesting that the uniformity constraint acts as a regularizer rather than guaranteeing perfect disentanglement, and the non-causal branch may still retain weak task-related cues. These results are consistent with our statement that MMCI effectively mitigates, rather than completely eliminates, spurious correlations.

\begin{table}[t]
\centering
\caption{Comparison of the number of parameters, training time, and inference time between MMCI and baselines on the CMU-MOSI dataset. All times are reported per epoch with batch size 8.}
\label{tab:param_count}
\begin{tabular}{l c c c}
    \toprule
    Method & Parameters & Training Time  & Inference Time \\
    \midrule
    EMOE~\cite{EMOE2025} & 317,295,771 & 15.99 s & 1.11 s \\
    GLoMo~\cite{GLoMo2024} & 109,818,887 & 13.37 s & 0.60 s \\
    C-MIB~\cite{MIB2023} & 190,930,180 & 16.09 s & 0.65 s \\
    ITHP~\cite{ITHP2024} & 184,883,706 & 21.16 s & 0.89 s \\
    \textbf{MMCI (Ours)} & 186,461,076 & 20.56 s & 0.71 s \\
    \bottomrule
\end{tabular}
\end{table}

\begin{figure}[t]
    \centering
    \includegraphics[width=0.49\linewidth]{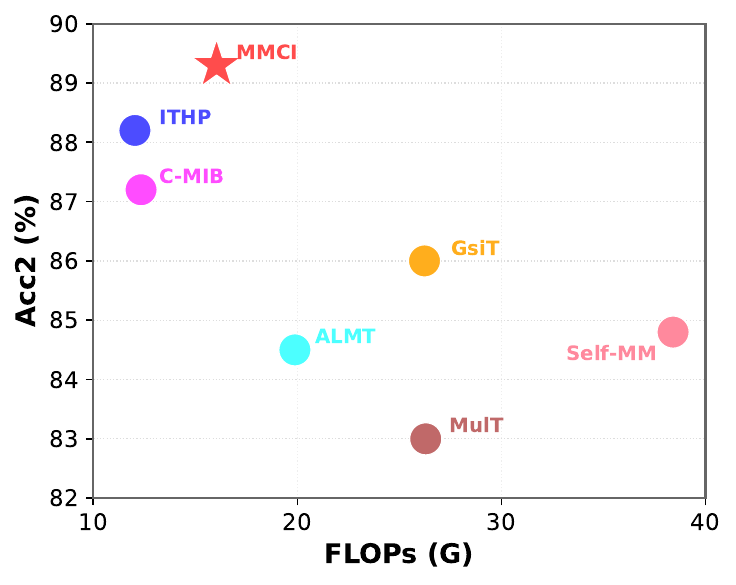}
    \hfill
    \includegraphics[width=0.49\linewidth]{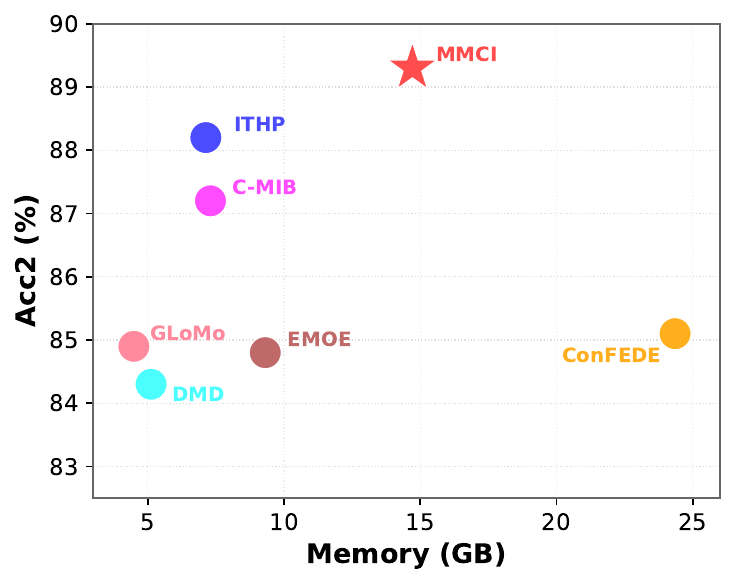}
    \caption{Comparison of FLOPs and peak GPU memory consumption. Peak memory is measured on CMU-MOSI with batch size 32.}
    \label{fig:efficiency}
\end{figure}

\subsection{Qualitative Analysis}
\label{sec:case_study}
To qualitatively investigate how MMCI mitigates spurious correlations, we select five representative test samples from the CMU-MOSI dataset, as shown in Fig.~\ref{fig:case_study}. We use ITHP~\cite{ITHP2024} as the baseline because it achieves the strongest overall performance among the compared conventional MSA methods. Case 1 examines potential inter-modal spurious correlations and the contributions of individual components, while Cases 2--3 and Cases 4--5 analyze intra-modal spurious correlations associated with visual attributes and lexical cues, respectively.

\subsubsection{Inter-modal spurious correlations and component analysis}
\begin{figure*}[t]
    \centering
    \includegraphics[width=\linewidth]{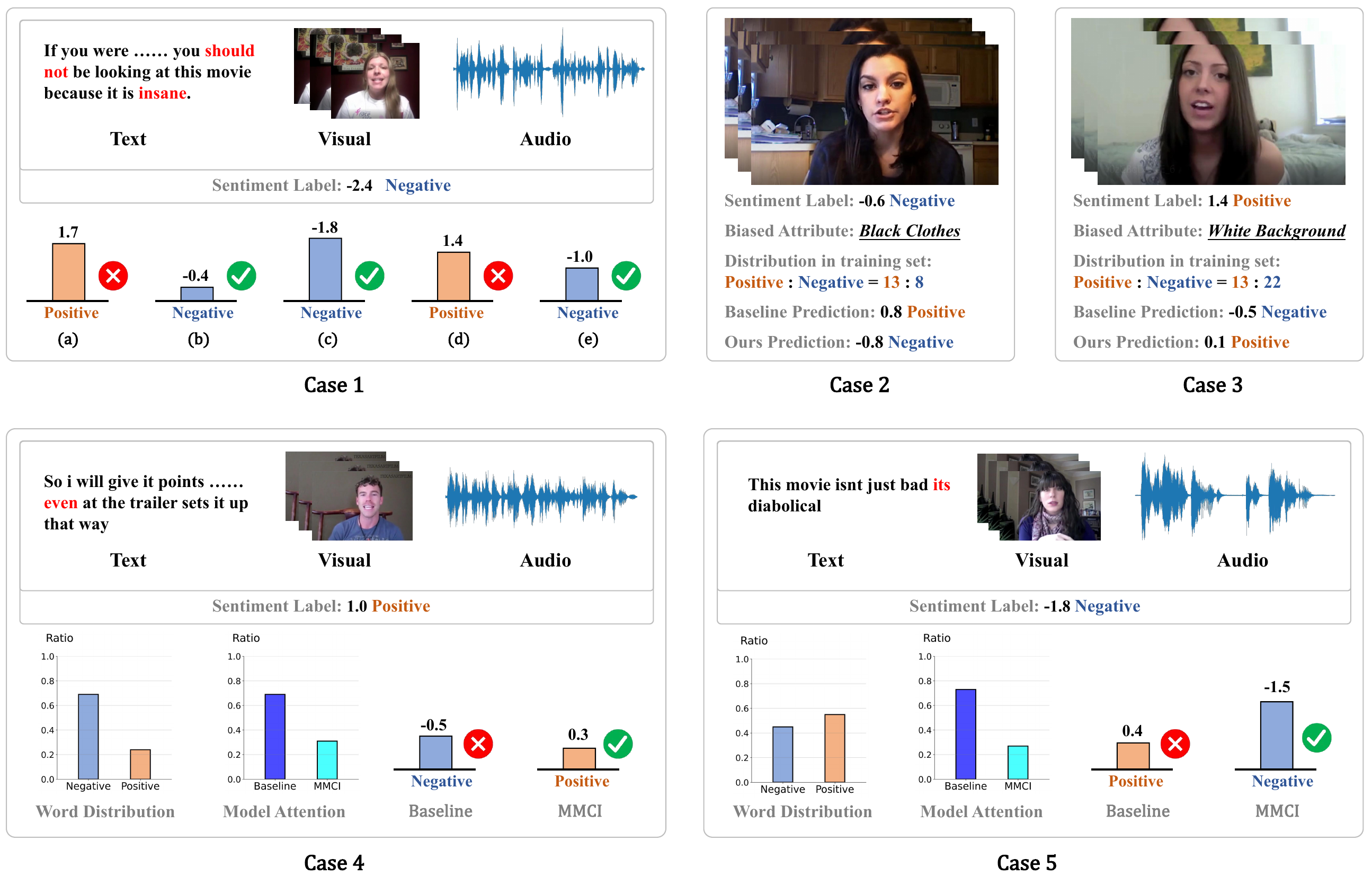}
    \caption{Qualitative case studies of MMCI on the CMU-MOSI dataset. For Case 1, (a)--(e) respectively show the predictions of ITHP, text-only ITHP, MMCI, MMCI without shortcut supervision, and MMCI without causal intervention. Cases 2 and 3 illustrate spurious correlations associated with visual attributes, while Cases 4 and 5 illustrate dataset-specific lexical associations. For Cases 4 and 5, the attention assigned by ITHP and MMCI to the selected biased word is normalized across the two models so that the reported ratios sum to one, representing relative inter-model attention rather than raw values.}
    \label{fig:case_study}
\end{figure*}

In Case 1, we revisit the test sample introduced in Fig.~\ref{fig:intro} and present the corresponding results in Fig.~\ref{fig:case_study}. The ground-truth sentiment score is $-2.4$. ITHP correctly predicts the sentiment polarity using only the textual modality but produces an incorrect prediction after incorporating the audio and visual modalities. This suggests that multimodal fusion may expose the model to inter-modal spurious correlations, causing unreliable cross-modal cues to override informative textual evidence. In contrast, MMCI predicts $-1.8$, which is substantially closer to the ground truth. Removing shortcut supervision changes the prediction to $1.4$, indicating that shortcut information remains in the learned representation. Removing shortcut-based intervention produces a less accurate prediction of $-1.0$. These observations provide qualitative evidence for the complementary contributions of causal--shortcut disentanglement and intervention.

\subsubsection{Intra-modal spurious correlations in the visual modality}
According to Sun et al.~\cite{GEAR2023}, the visual shortcut attributes in Cases 2 and 3 are black clothing and the white background, respectively. In Case 2, black clothing occurs in 13 positive and 8 negative training videos. ITHP follows this superficial positive association and predicts the wrong polarity, whereas MMCI relies more on sentiment-relevant cues, such as the speaker's serious expression, and produces the correct prediction. Similarly, in Case 3, ITHP is misled by the spurious association of the white background, whereas MMCI correctly identifies the sentiment polarity. These cases demonstrate that MMCI can effectively mitigate this type of visual spurious correlation.

\subsubsection{Intra-modal spurious correlations in the textual modality}
Cases 4 and 5 examine sentiment-neutral words with biased training-label distributions. In Case 4, ``even'' exhibits a strong association with negative sentiment in the training set, although the positive sentiment in the current sentence is conveyed by ``give it points.'' ITHP follows this spurious negative association and assigns high attention to ``even,'' resulting in an incorrect sentiment polarity. In contrast, the causal branch of MMCI assigns substantially less attention to this lexical shortcut and produces the correct prediction. Similarly, in Case 5, ITHP is misled by the sentiment-neutral word ``its,'' whereas MMCI assigns less causal attention to it and focuses more on the sentiment-bearing words ``bad'' and ``diabolical.'' These cases demonstrate that MMCI can effectively mitigate spurious correlations associated with dataset-specific lexical shortcuts.

\subsection{Model Complexity and Computational Efficiency Analysis} 
We evaluate the computational efficiency of MMCI from four perspectives: parameter size, training and inference time, FLOPs, and peak GPU memory. As shown in Table~\ref{tab:param_count}, MMCI does not introduce significant additional computational cost in terms of parameter size, runtime, and FLOPs compared with state-of-the-art methods. Specifically, MMCI has only a marginal increase in parameters compared with ITHP~\cite{ITHP2024} (less than 1\%), while remaining more lightweight than EMOE~\cite{EMOE2025} and C-MIB~\cite{MIB2023}. In terms of training and inference time, MMCI is faster than ITHP, demonstrating its favorable runtime efficiency. For FLOPs, as shown in Fig.~\ref{fig:efficiency}, MMCI is only slightly higher than information bottleneck-based methods such as ITHP and C-MIB, but remains significantly lower than methods based on hierarchical modal-wise heterogeneous graphs, such as GsiT~\cite{jin2025multimodal}. In terms of memory consumption, MMCI is relatively higher than most baselines because its relation-specific GATs require multiple adjacency matrices for intra-modal and inter-modal relations. Letting \(n=3L\) and \(|\mathcal{R}|=6\), dense graph attention requires \(\mathcal{O}(|\mathcal{R}|n^{2}d)\) computation and \(\mathcal{O}(|\mathcal{R}|n^{2})\) storage, explaining the increased peak memory usage. The causal intervention only introduces batch-level shortcut recombination during training and is omitted during inference, which relies on the causal branch alone. For large-scale datasets, sparse adjacency storage and neighborhood pruning can further reduce this overhead. Overall, MMCI maintains a favorable balance between efficiency and effectiveness.

\section{Conclusion}\label{sec:conclusion}
In this work, we investigate multimodal fusion in MSA from a causal perspective and propose a Multi-relational Multimodal Causal Intervention (MMCI) framework to mitigate spurious correlations within and across modalities. MMCI constructs a multi-relational graph to model intra- and inter-modal dependencies, employs complementary attention to separate causal and shortcut representations, and approximates backdoor adjustment by evaluating causal features under stratified shortcut contexts. Extensive experiments demonstrate that MMCI consistently achieves superior performance in both standard and distribution-shifted settings. The ablation studies and qualitative case analyses further confirm the contributions of relation modeling, causal--shortcut disentanglement, and shortcut-based intervention.
Beyond MSA, the core idea of MMCI, which encourages learning stable causal relations while reducing reliance on unstable shortcut correlations, may also be applicable to other multimodal tasks. By adapting the relation definitions and task-specific prediction heads, MMCI could be extended to multimodal healthcare analysis, emotion recognition, human--computer interaction, and other multimodal reasoning tasks, where reliable decision-making similarly depends on stable evidence across heterogeneous modalities rather than dataset-specific shortcuts.

However, MMCI has several limitations. First, the uniformity constraint suppresses label-discriminative information that can be decoded by the auxiliary shortcut predictor but cannot guarantee perfect causal disentanglement; when causal and shortcut factors are highly entangled, it may also suppress useful complementary information. Second, because latent confounders are unobservable, the shortcut memory bank provides only a finite empirical approximation to the stratification required by backdoor adjustment and may not cover all possible shortcut contexts. Third, although the adopted datasets are standard MSA benchmarks, their scale and source diversity remain limited. CMU-MOSI and CMU-MOSEI are mainly collected from online videos and may contain speaker-, topic-, platform-, and recording-specific patterns, while CH-SIMS is derived primarily from movies and television programs and may include scripted expressions and production-related biases. These dataset-specific characteristics can induce spurious correlations between sentiment labels and lexical, acoustic, or visual cues, limiting generalization to other populations, domains, languages, and recording environments. Moreover, the OOD version of CMU-MOSI mainly evaluates shifts in word--sentiment correlations and cannot cover all possible real-world distribution shifts. Fourth, maintaining multiple relation-specific causal and shortcut representations results in relatively high peak GPU memory consumption. Future work will explore more adaptive shortcut modeling, larger and more diverse cross-domain datasets, and lightweight, memory-efficient implementations.

\bibliographystyle{IEEEtran}
\bibliography{IEEEexample}


\end{document}